\documentclass[lettersize,journal]{IEEEtran}
\usepackage{mathtools}
\usepackage{color,soul}
\usepackage{cite}
\usepackage{titlesec}
\usepackage{amsmath}
\usepackage{makecell}
\usepackage{amsfonts}
\usepackage{hyperref}
\usepackage{algpseudocode}
\usepackage{algorithm}
\usepackage{cleveref}
\crefname{assumption}{assumption}{assumptions}
\usepackage{arydshln}
\usepackage{caption}
\usepackage{balance} 
\usepackage{xcolor}
\usepackage{subcaption}
\usepackage{graphicx}
\usepackage{amssymb}
\usepackage{textcomp}
\usepackage{multirow}
\usepackage{kotex}
\usepackage{changes}
\usepackage{amsthm}

\newtheoremstyle{case}{}{}{}{}{}{:}{ }{}
\theoremstyle{case}

\usepackage{siunitx}
\DeclareMathOperator*{\argmin}{arg\,min}

\usepackage{breqn}
\usepackage{gensymb}
\usepackage{mathtools}
\begin{document}
\title{Efficient COLREGs-Compliant Collision Avoidance using Turning Circle-based Control Barrier Function}
\author{Changyu Lee, Jinwook Park, and Jinwhan Kim
\thanks{Manuscript received April 10, 2000; revised August 16, 2000; accepted September 24, 2000. Date of publication October 20, 2020; date of current version March 9, 2022. \it{(Corresponding author: Jinwhan Kim.)}}
\thanks{C. Lee, J. Park, and J. Kim are with the Mobile Robotics and Intelligence Laboratory, Department of Mechanical Engineering, Korea Advanced Institute of Science and Technology (KAIST), Daejeon 34141, South Korea. (e-mails: leeck@kaist.ac.kr, qkrwlsdnr10@kaist.ac.kr, jinwhan@kaist.ac.kr)}}
\markboth{Journal of \LaTeX\ Class Files,~Vol.~14, No.~8, May~2025}%
{Shell \MakeLowercase{\textit{et al.}}: A Sample Article Using IEEEtran.cls for IEEE Journals}

\maketitle

\begin{abstract}

This paper proposes a computationally efficient collision avoidance algorithm using turning circle-based control barrier functions (CBFs) that comply with international regulations for preventing collisions at sea (COLREGs). 
Conventional CBFs often lack explicit consideration of turning capabilities and avoidance direction, which are key elements in developing a COLREGs-compliant collision avoidance algorithm. To overcome these limitations, we introduce two CBFs derived from left and right turning circles.
These functions establish safety conditions based on the proximity between the traffic ships and the centers of the turning circles, effectively determining both avoidance directions and turning capabilities.
The proposed method formulates a quadratic programming problem with the CBFs as constraints, ensuring safe navigation without relying on computationally intensive trajectory optimization.
This approach significantly reduces computational effort while maintaining performance comparable to model predictive control-based methods.
Simulation results validate the effectiveness of the proposed algorithm in enabling COLREGs-compliant, safe navigation, demonstrating its potential for reliable and efficient operation in complex maritime environments.

\end{abstract}

\begin{IEEEkeywords}
Control barrier function, autonomous ship, COLREGs
\end{IEEEkeywords}

\section{Introduction} \label{sec:intro}
Recently, autonomous ships have attracted significant attention for their potential applications in transportation, environmental monitoring, and search and rescue operations. Their use helps prevent human errors, which are the leading cause of maritime accidents, thereby reducing ship-related incidents and lowering maintenance costs.

However, despite these advantages, the deployment of autonomous ships in maritime environments remains limited. The main challenge is developing technologies that ensure safety in mixed traffic conditions, where autonomous and human-operated ships coexist. To address this, it is essential to develop algorithms that comply with the international regulations for preventing collisions at sea (COLREGs), which serve as maritime traffic rules. These regulations dictate appropriate evasive maneuvers for various encounter scenarios, and ensuring adherence to them is essential for the safe and efficient integration of autonomous ships into maritime operations.

The velocity obstacle (VO) method is a simple and widely used approach in collision avoidance. In this method, the VO assumes that both the own ship and the traffic ship maintain their current velocities. The VO defines the set of all velocities that would lead to a collision, and the own ship's velocity is constrained to ensure it stays outside of this set. Additionally, a forbidden zone can be incorporated into the VO to consider the COLREGs regulations to ensure compliance with maritime navigation rules \cite{vo2_thyri2022partly, vo4_zhang2021colregs, vo5_shaobo2020collision, vo6_cho2019experimental, vo7_zheng2023regulation, vo8_huang2018velocity}.
The VO method has been integrated with sampling-based path planning algorithms, such as the VO-RRT approach \cite{vorrt_dubey2021vorrt}. 
Several variations have been introduced to improve the performance and applicability of the VO approach.
The generalized VO (GVO) method considers the dynamic behavior of ships \cite{gvo_huang2019generalized}, while the probabilistic VO (PVO) method incorporates uncertainty in vehicle motion prediction \cite{pvo_cho2020efficient}. To address the oscillation problem often observed in VO-based navigation, the reciprocal VO (RVO) method has been introduced \cite{drvo_kufoalor2018proactive}.

To achieve better performance, model predictive control (MPC)-based approaches have gained increasing adoption in recent years. MPC solves an optimization problem at each time step, formulating the control problem over a future time horizon while considering input and state constraints. Due to its ability to explicitly handle constraints and predict future vehicle behavior, MPC has been shown to outperform conventional guidance and tracking control methods in autonomous ship applications \cite{mpc1_abdelaal2018nonlinear, mpc9_lee2023nonlinear, mpc10_kim2023navigable, mpc3_du2022multi, mpc11_he2023model}.
Earlier studies focused on sampling-based MPC methods that incorporate COLREGs, with experimental validations confirming their effectiveness \cite{smpc1_johansen2016ship, smpc2_hagen2018mpc, smpc3_kufoalor2019autonomous, scmpc1_tengesdal2021ship, scmpc2_hagen2022scenario, scmpc3_akdaug2022collaborative, scmpc4_tengesdal2022ship}. Additionally, optimization-based MPC approaches have been widely studied, with COLREGs-aware MPC methods introduced to enhance maritime navigation \cite{mpc2_eriksen2020hybrid, mpc5_tsolakis2022colregs, mpc6_tsolakis2024model, mpc8_breivik2017mpc}. These methods have been applied in constrained environments, such as canals \cite{mpc4_de2022regulations}, and for multi-vessel coordination scenarios \cite{mpc7_du2021mpc}. However, MPC-based algorithms often require substantial computational resources, which can limit their practicality in real-time applications.

Control barrier functions (CBFs) ensure safety by guaranteeing the forward invariance of a defined safe set \cite{ames2016control, ames2019control}. This method enforces safety constraints, which are expressed as state-dependent conditions, by restricting the control input to ensure the vehicle continuously satisfies these constraints.
A quadratic programming (QP) problem can be formulated based on CBFs to create a safety filter, which minimally modifies the control inputs from any nominal controller to enforce CBF constraints. Owing to its relatively simple formulation, this approach is more computationally efficient than complex trajectory optimization-based algorithms and has been widely adopted, either as a standalone solution or in combination with trajectory optimization.
Recently, CBFs have been applied in various safety-critical applications \cite{zeng2021safety}, such as drone geofencing \cite{singletary2022onboard}, quadruped locomotion \cite{unlu2024control}, and aircraft collision avoidance \cite{molnar2025collision}. In the maritime domain, CBFs have been applied to ship collision avoidance \cite{cbf1_xue2024human, cbf2_xu2022multi, cbf4_xu2024safety, cbf6_basso2020safety, fan2024wake}, and multi-ship interaction scenarios \cite{cbf7_xu2022multi, cbf8_wu2024constrained}.

While CBFs have the potential to provide an effective framework for COLREGs-compliant collision avoidance, their application in this domain has received relatively limited attention.
One of the main challenges is that conventional CBFs lack explicit mechanisms for determining avoidance directions, which is a critical requirement for complying with COLREGs, and do not adequately account for the turning capabilities of ships.
A common approach in CBF-based obstacle avoidance is to maintain a predefined safe distance from obstacles, using the equation $h = d^2-r^2$ \cite{vulcano2022safe, jin2022collision, manjunath2021safe, peng2023safety, bruggemann2022simultaneous, jian2023dynamic}, where $d$ is the Euclidean distance to the traffic ship and $r$ is the safety margin. However, this method does not consider the turning capabilities of ships, and it is difficult to enforce avoidance in a specific direction. 
To address this, a COLREGs-compliant CBF-based method was proposed in \cite{cbf3_thyri2020reactive, thyri2022domain}, which combines CBFs with a target ship domain algorithm to determine avoidance direction. However, this approach does not account for the nonholonomic constraints of ships. In \cite{fan2024wake}, the ship trajectory was parameterized using circular motion to constrain control input and smooth the avoidance path. While this method improves motion smoothness, it still has limited capability to determine appropriate avoidance directions and does not fully account for the maximum turning capabilities of ships. To address turning capabilities, turning circle-based CBFs (TC-CBFs) were introduced \cite{arxiv-tccbf}, which explicitly consider the maximum turning capabilities of ships. However, this approach is also limited in its ability to autonomously determine appropriate avoidance directions in compliance with COLREGs.

To overcome these limitations, we use two distinct TC-CBFs: the left TC-CBF (LTC-CBF) and the right TC-CBF (RTC-CBF) for COLREGs-compliant collision avoidance in autonomous ships. These are constructed based on the current heading, speed, and maximum turning rate for each direction, thus forming left and right turning circles.  
The safety condition is defined by the proximity between the traffic ship and the relevant turning circle, ensuring that there remains sufficient maneuvering space to avoid collisions in either direction using the ship’s maximum turning capability. This formulation allows the ships to determine an appropriate avoidance direction while accounting for the turning limitations, thereby enabling effective integration with COLREGs rules.
Compared to existing work \cite{arxiv-tccbf}, which combines LTC-CBF and RTC-CBF within an MPC framework to enhance obstacle avoidance efficiency for nonholonomic vehicles, this paper uses the two CBFs independently to enforce COLREGs rules explicitly. In addition, the proposed method is extended to handle multi-obstacle collision avoidance scenarios. 

Finally, a QP-based safety filter is designed using the proposed CBF constraints, modifying the nominal control input to ensure both safety and rule compliance. 
The effectiveness of the proposed method is demonstrated through comparison with a state-of-the-art MPC framework, which enforces state constraints to comply with COLREGs rules. Numerical simulations with various scenarios demonstrate that our method achieves significantly higher computational efficiency while producing collision avoidance performance comparable to MPC-based approaches.

The main contributions of this study are summarized as follows:
\begin{itemize}
\item We introduce two TC-CBF approaches, LTC-CBF and RTC-CBF, for COLREGs-compliant collision avoidance. These functions determine the avoidance direction while considering the turning capabilities of the ships.
\item The proposed method is validated through numerical simulations and compared with an MPC-based collision avoidance framework, demonstrating a high computational efficiency while maintaining comparable performance.
\item By using a CBF-QP-based safety filter framework, the proposed approach can be integrated with any nominal controller, making it highly applicable for real-world scenarios.
\end{itemize}

This paper is organized as follows. The following section gives preliminaries, and the decision-making module is presented in Section~\ref{sec:decision_making}. In Section~\ref{sec:cbf_colav}, the proposed CBF-based COLREGs-compliant collision avoidance algorithm is formulated. Section~\ref{sec:simulation} shows the simulation results, and Section~\ref{sec:conclusion} concludes this paper.

\section{Preliminaries} \label{sec:pre}
% \subsection{Notations}
% The notation $||x||^{2}_{W} $ represents $x^\top W x$. 
% Let $\mathit{I}_{n\times m}$ and $\mathbf{0}_{n\times m}$ be the $n \times m$ identity matrix and zero matrix, respectively. If $n=m$, we use the shorthand subscript $(\cdot)_n$ instead of $(\cdot)_{n\times m}$.
% The subscript $(\cdot)_{k|t}$ represents the value at time $t+k$, which is predicted at time $t$.

\subsection{International Regulations for Preventing Collisions at Sea (COLREGs)}

\begin{figure}[t]
    \captionsetup[subfigure]{justification=centering}
        \centering
    \begin{subfigure}[h]{0.32\linewidth}
        \centering
        \includegraphics[width=\textwidth]{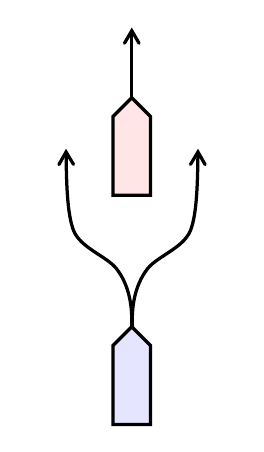}
        \caption{Overtaking.}
        \label{fig:colregs_rule_overtaking}
    \end{subfigure}    
    \begin{subfigure}[h]{0.32\linewidth}
        \centering
        \includegraphics[width=\textwidth]{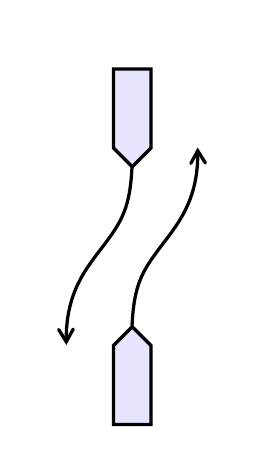}
        \caption{Head-on.}
        \label{fig:colregs_rule_headon}
    \end{subfigure}
    \begin{subfigure}[h]{0.32\linewidth}
        \centering
        \includegraphics[width=\textwidth]{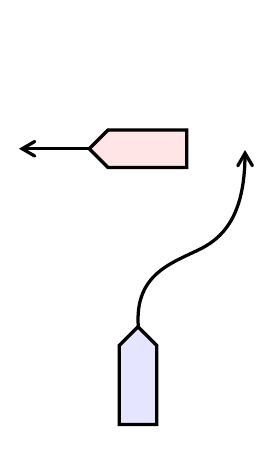}
        \caption{Crossing.}
        \label{fig:colregs_rule_crossing}
    \end{subfigure}
    \caption{COLREGs-compliant evasive maneuvers. Blue ships take an evasive maneuver (give-way), while red ships maintain their course and speed (stand-on) in accordance with COLREGs.}
    \label{fig:colregs_rule}
\end{figure}

\begin{figure*}[t]   
\centerline{\includegraphics[width=\linewidth]{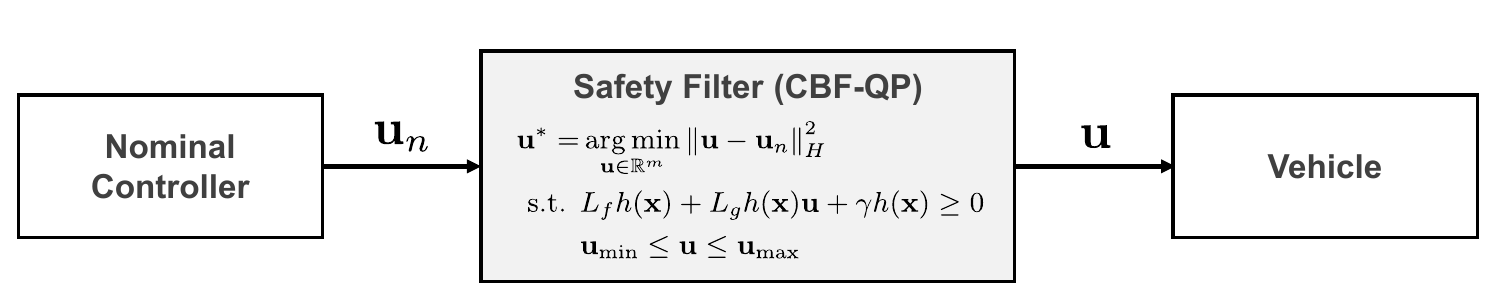}}
    \caption{Architecture of the CBF-QP safety filter.}
    \label{fig:cbf_qp}
\end{figure*}

The COLREGs, established by the International Maritime Organization (IMO), provide the primary legal framework for safe and standardized maritime navigation worldwide. These regulations define the responsibilities of ships in various navigational scenarios. Among the COLREGs, Rules 13-17 are particularly critical for decision-making in collision avoidance systems. As stated in \cite{COLREG}, Rule 13 specifies that when one vessel is overtaking another, the overtaking vessel must keep out of the way. Rule 14, also known as the head-on rule, states that when two power-driven vessels are approaching each other head-on, both should alter course to starboard.
Rule 15 addresses crossing situations, stating that when two power-driven vessels are crossing paths, the vessel that has the other on its starboard side must give way.
Rule 16 states that a vessel required to keep out of the way of another vessel should take early and substantial action to avoid a collision. Rule 17 is a rule of the road that describes how a stand-on vessel should act to avoid a collision with a give-way vessel. 
These rules serve as the foundation for autonomous collision avoidance algorithms, influencing path-planning, decision-making, and control systems. Three representative scenarios governed by these rules are illustrated in Fig.~\ref{fig:colregs_rule}.
Ensuring that an autonomous ship correctly interprets and applies these regulations is essential for achieving safe and predictable navigation in maritime environments.

\subsection{Control Barrier Functions} \label{sec:subsec}
Consider a nonlinear control-affine system:
\begin{equation} \label{eq:system}
    \dot{\mathbf{x}} = f(\mathbf{x}) + g(\mathbf{x})\mathbf{u},
\end{equation}
where $\mathbf{x}\in \mathcal{X}\subset \mathbb{R}^n$ and $\mathbf{u}\in \mathcal{U} \subset \mathbb{R}^m$ represent the state and control input vectors, respectively.
The set $\mathcal{S} \subset \mathbb{R}^n$ defines the safe region within which the vehicle's states must remain. The objective is to design a controller that guarantees the forward invariance of the set $\mathcal{S}$. This ensures that if $\mathbf{x}(0) \in \mathcal{S}$, then $\mathbf{x}(t) \in \mathcal{S}$ for all $t \geq 0$, ensuring the vehicle consistently remains in a safe state. Forward invariance of the safe set can be achieved through the use of a CBF \cite{ames2016control}.

Let $\mathcal{S}$ be defined as the superlevel set of a continuously differentiable function $h: \mathbb{R}^n \rightarrow \mathbb{R}$, formulated as:
\begin{equation}
\begin{aligned}
\mathcal{S} & =\left\{\mathbf{x} \in \mathbb{R}^n: h(\mathbf{x}) \geq 0\right\}, \\
\partial \mathcal{S} & =\left\{\mathbf{x} \in \mathbb{R}^n: h(\mathbf{x})=0\right\}, \\
\operatorname{Int}(\mathcal{S}) & =\left\{\mathbf{x} \in \mathbb{R}^n: h(\mathbf{x})>0\right\} .
\end{aligned}
\end{equation}
The function $h$ is a CBF if $ \textstyle \frac{\partial h}{\partial \mathbf{x}}\mathbf{x} \neq 0 $ for all $\mathbf{x} \in \partial \mathcal{S}$, and there exists an extended class-$\mathcal{K}$ function\footnote{A function $\gamma: (-b, a) \rightarrow (-\infty, \infty)$ is of extended class-$\mathcal{K}$ if it is strictly increasing and satisfies $\gamma(0) = 0$, for $a, b > 0$.} $\alpha$ such that for the system \eqref{eq:system} and for all $\mathbf{x} \in \mathcal{S}$ \cite{ames2019control}:
\begin{equation} \label{eq:cbfconstraints}
\sup _{\mathbf{u} \in \mathcal{U}}[L_f h(\mathbf{x})+L_g h(\mathbf{x}) \mathbf{u}+{\alpha}(h(\mathbf{x}))] \geq 0, 
\end{equation}
where $\textstyle L_f h(\mathbf{x})=\frac{\partial h(\mathbf{x})}{\partial \mathbf{x}} f(\mathbf{x})$ and $\textstyle L_g h(\mathbf{x})=\frac{\partial h(\mathbf{x})}{\partial \mathbf{x}} g(\mathbf{x})$ denote the Lie derivatives of $h(\mathbf{x})$ along $f$ and $g$, respectively.
The admissible control space $U(\mathbf{x})$ is defined as:
\begin{equation}
U(\mathbf{x})=\left\{\mathbf{u} \in \mathcal{U} : L_f h(\mathbf{x})+L_g h(\mathbf{x}) \mathbf{u}+\alpha(h(\mathbf{x})) \geq 0\right\},
\end{equation}
ensuring the forward invariance of $\mathcal{S}$.

The choice of the function $\alpha$ influences how the state approaches the boundary of $\mathcal{S}$. 
A common choice for $\alpha(h)$ is a scalar multiple of $h(\mathbf{x})$, typically $\alpha(h(\mathbf{x}))= \gamma h(\mathbf{x})$ with $\gamma>0$.
This condition requires the control input to satisfy:
\begin{equation} 
\label{eq:cbf_ic}
L_f h(\mathbf{x})+L_g h(\mathbf{x}) \mathbf{u}+\gamma h(\mathbf{x}) \geq 0.
\end{equation}

\subsection{CBF-based Safety Filters using Quadratic Programming}
A CBF-based safety filter ensures that a system remains within a predefined safe region while achieving its control objectives. It modifies the nominal control input to satisfy CBF constraints while minimizing deviation from the desired performance. CBF-based safety filters are typically implemented using the following QP formulation (CBF-QP):
\begin{equation}
\begin{aligned}
 \mathbf{u}^* = & \argmin_{\mathbf{u}\in\mathbb{R}^m}  \left\| \mathbf{u}-\mathbf{u}_{n} \right\|^2_H \\
\text { s.t. } & L_f h(\mathbf{x})+L_g h(\mathbf{x}) \mathbf{u}+\gamma h(\mathbf{x}) \geq 0, \\
& \mathbf{u}_{\text{min}} \leq \mathbf{u} \leq \mathbf{u}_{\text{max}},
\end{aligned}
\label{eq:qp_cbf}
\end{equation}
where the notation $||\cdot||^{2}_{W} $ represents $(\cdot)^\top W (\cdot)$, $H$ is a positive definite weighting matrix, and the last inequality constraints represent input saturation limits of the vehicle. The QP in \eqref{eq:qp_cbf} ensures the satisfaction of safety constraints while keeping the control input as close as possible to the nominal control input $\mathbf{u}_n$. The CBF-QP framework is illustrated in Fig.~\ref{fig:cbf_qp}.
These filters operate efficiently in real-time, dynamically adapting to changing environments, and are particularly effective for applications in robotics \cite{jankovic2023multiagent} and autonomous vehicles \cite{cbf2_xu2022multi, arxiv-tccbf}. 

\section{Encounter Situation Decision Making} \label{sec:decision_making}
\begin{figure}[t!]
\centerline{\includegraphics[width=\linewidth]{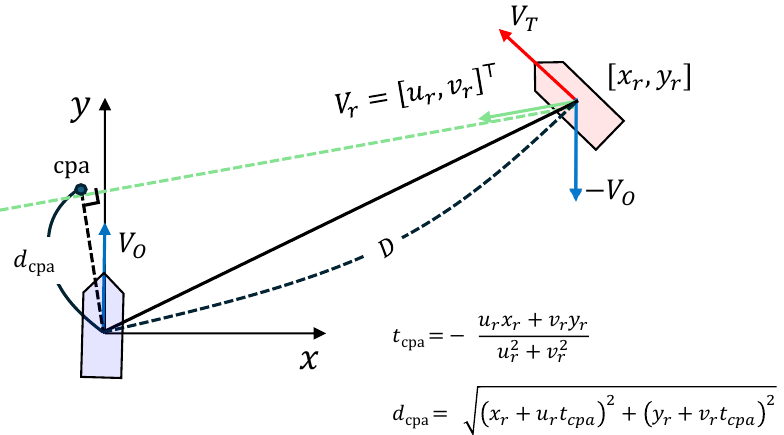}}
    \caption{Illustration of the DCPA and TCPA.}
    \label{fig:DCPA_TCPA}
\end{figure}

In this chapter, we present the encounter situation decision-making module, which is crucial for determining which ships must be considered in the collision avoidance module, as well as identifying the type of encounter according to COLREGs rules. For this purpose, concepts based on the closest point of approach (CPA) are widely used and also adopted in this study.
When a traffic ship moves with a relative velocity $V_{r}$ with respect to the own ship, the point at which the perpendicular distance from the own ship to the line defined by the direction of $V_{r}$ is minimized is referred to as the CPA. The corresponding distance is known as the distance at CPA (DCPA), and the time required for the traffic ship to reach this point is called the time to CPA (TCPA), as illustrated in Fig.~\ref{fig:DCPA_TCPA}. Based on these concepts, a traffic ship is classified as an encountered ship if both the TCPA and DCPA are below predefined thresholds or if the current distance between the ships falls below a minimum distance threshold.

\begin{figure}[t]
    \centerline{\includegraphics[width=\linewidth]{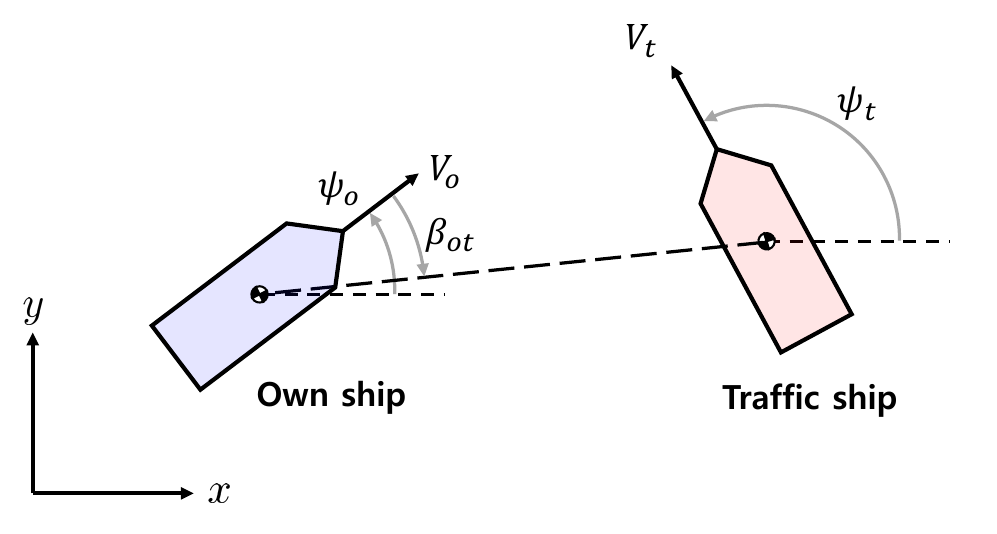}}
    \caption{Coordinate systems used for encounter classification.}
    \label{fig:coordinate}
\end{figure}

\begin{figure}[t]
    \captionsetup[subfigure]{justification=centering}
        \centering
    \begin{subfigure}[h]{\linewidth}
        \centering
        \includegraphics[width=.95\textwidth]{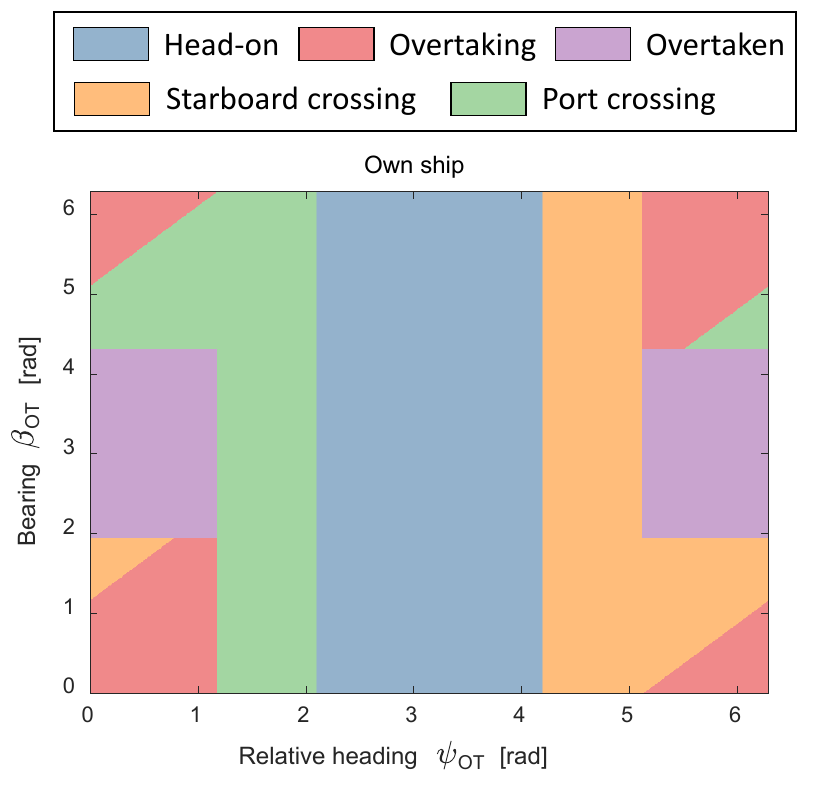}
        \vspace{-0.3cm}
        \caption{Own ship.}
        \label{fig:COLREGs_OS}
    \end{subfigure}
    \begin{subfigure}[h]{\linewidth}
        \vspace{0.15cm}
        \centering
        \includegraphics[width=.95\textwidth]{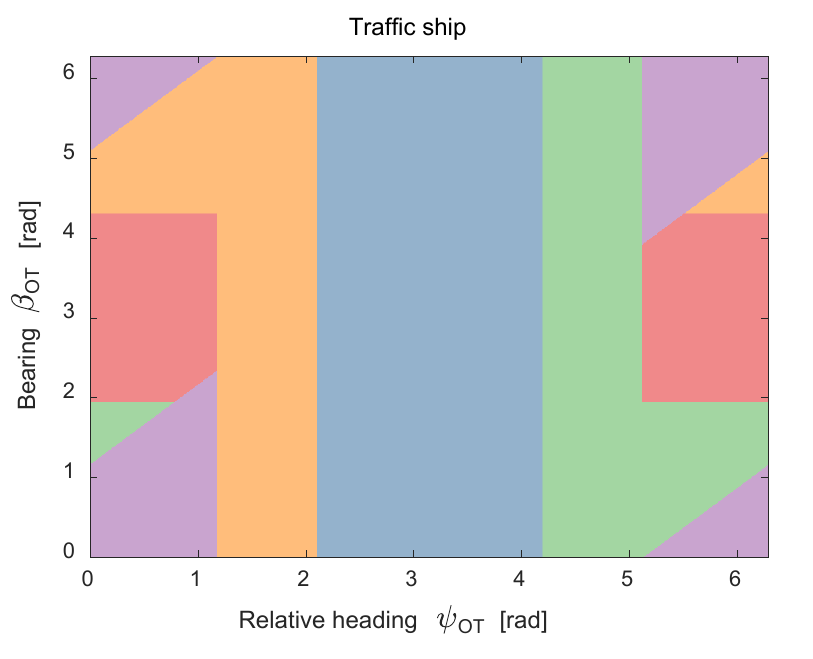}
        \vspace{-0.3cm}
        \caption{Traffic ship.}
        \label{fig:COLREGs_TS}
    \end{subfigure}    
    \caption{Illustration of COLREGs encounter types.}
    \label{fig:COLREGs_map}
\end{figure}

Once a traffic ship is identified as an encountered ship, it must be further classified according to the relevant COLREGs rule.
The encounter type is determined based on the bearing $\beta_{ot}$ and the relative heading $\psi_{ot} = \psi_t - \psi_o$ between the own ship and the traffic ship, as illustrated in Fig.~\ref{fig:coordinate}. 
This study considers the following COLREGs encounter types: head-on, starboard crossing, port crossing, overtaking, and overtaken. Head-on and crossing scenarios are classified based on the relative heading, whereas overtaking and overtaken cases are identified using the relative bearing.
To ensure consistency in classification between the own ship and the traffic ship, symmetric interpretation is essential. 
For example, if the own ship is classified as head-on, the traffic ship must also be classified as head-on. Similarly, if the own ship is in a starboard crossing situation, the traffic ship must be in a port crossing situation, and vice versa. In the case of overtaking, if the own ship is overtaking, the traffic ship must be classified as overtaken, and vice versa.
Accordingly, this study adopts the symmetry-guaranteed COLREGs classification method proposed in \cite{pvo_cho2020efficient}. Let $(x_o, y_o)$ and $(x_t, y_t)$ denote the position of the own ship and the traffic ship, respectively. The relative heading $\psi_{ot}$ and relative bearing $\beta_{ot}$ are computed as follows:
\begin{equation}
\begin{aligned}
\psi_{ot} &= \psi_t - \psi_o, \\ 
\beta_{ot} &= \text{atan2}(y_t - y_o, x_t - x_o)-\psi_o.     
\end{aligned}
\end{equation}
Based on the relative heading and bearing, the classification results for both the own ship and the target ship can be obtained, as shown in Fig.~\ref{fig:COLREGs_map}. (For further details, refer to Algorithm 1 in \cite{pvo_cho2020efficient}.)

\section{COLREGs-Compliant Collision Avoidance using Control Barrier Function} \label{sec:cbf_colav}
\subsection{Guidance and Control Algorithms for Waypoint Following}
In COLREGs-compliant collision avoidance, precise control is not required. The primary focus is on ensuring safe and rule-compliant maneuvers rather than exact path following. For this purpose, the unicycle model is sufficient as it captures the fundamental motion constraints of nonholonomic ships. The model is defined as follows:
\begin{equation}  
\begin{aligned}
    \dot{x} &= u\cos\psi, \\ 
    \dot{y} &= u\sin\psi, \\
    \dot{\psi} &= r, \\
    \dot{u} &= a,  
    \label{unicycle}
    \end{aligned}
\end{equation}
where $\mathbf{u} = [a,r]^\top \in \mathbb{U}$ represents the control inputs, with $a$ being the forward acceleration and $r$ the turning rate. The variables $x$, $y$, and $\psi$ denote the vehicle's position and heading, respectively.

To generate the nominal control input to follow the desired waypoint path, this study adopts the LOS guidance and PD tracking control algorithms \cite{lekkas2013line}. 
To achieve both velocity and heading control, we design a PD controller where the control inputs are acceleration $a$ and angular velocity $r$. The velocity control law is given by the speed error $e_u = u_d - u$, as follows:
\begin{equation}
\label{eq:acc_pd}
a_n = K_{p}^u e_u  + K_{d}^u \dot{e}_u,
\end{equation}
where $u$ and $u_d$ represent the current and desired speed, respectively, and $\dot{e}_u$ denotes error time derivative of the speed error. The proportional gain $K_{p}^u$ and the derivative gain $K_{d}^u$ regulate the tracking performance.
Similarly, the heading control law is formulated with the heading error $e_\psi = \psi_d - \psi$ as follows:
\begin{equation}
\label{eq:rot_pd}
r_n = K_p^\psi e_\psi + K_d^\psi \dot{e}_\psi,
\end{equation}
where $\psi$ and $\psi_d$ are the current and desired heading angles, respectively, and $\dot{e}_\psi$ corresponds to the time derivative of the heading error. The control gains $K_p^\psi$ and $K_d^\psi$ are selected to ensure stable and responsive heading tracking.

\begin{figure}[t]
    \captionsetup[subfigure]{justification=centering}
        \centering
    \begin{subfigure}[h]{\linewidth}
        \centering
        \includegraphics[width=.9\textwidth]{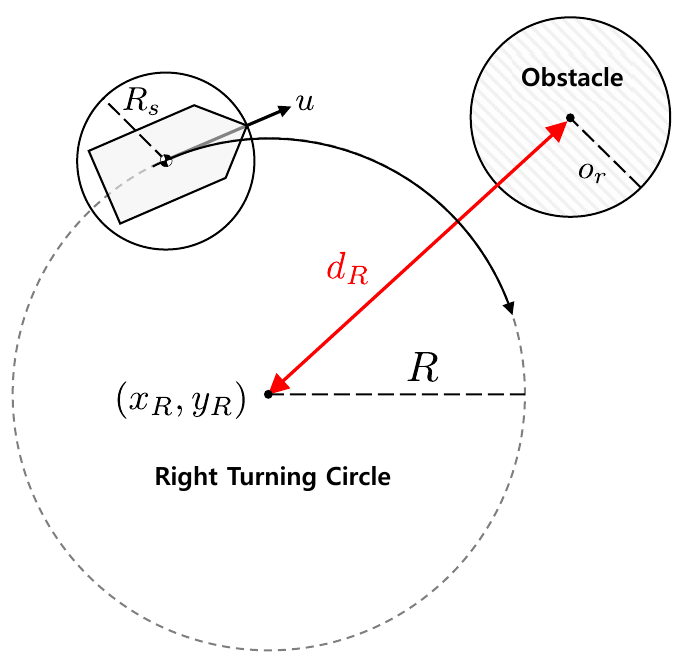}
        \caption{Right turning circle.}
        \label{fig:rtc}
        \vspace{0.3cm}
    \end{subfigure}
    \begin{subfigure}[h]{\linewidth}
        \centering
        \includegraphics[width=\textwidth]{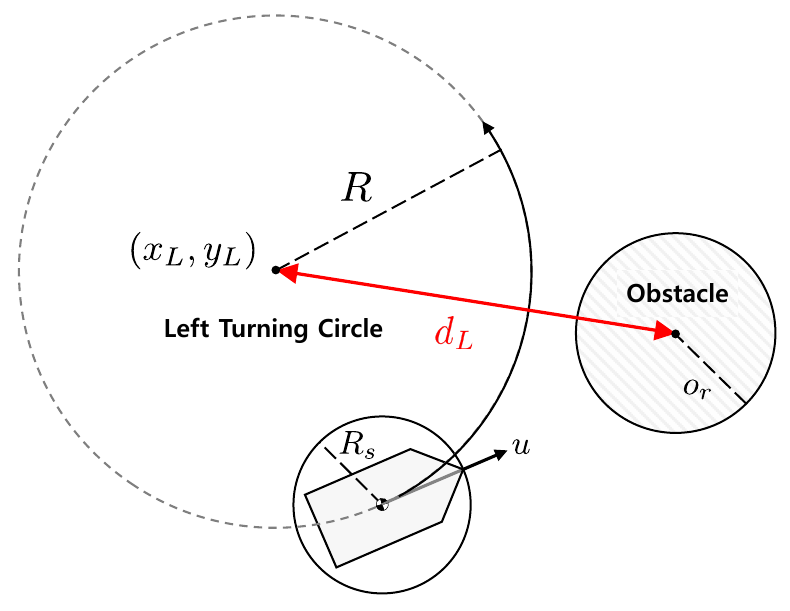}
        \caption{Left turning circle.}
        \label{fig:ltc}
    \end{subfigure}    
    \caption{Geometrical representation of left and right turning circles.}
    \label{fig:turning_circles}
\end{figure}

\begin{figure*}[t]
    \captionsetup[subfigure]{justification=centering}
        \centering
    \begin{subfigure}[h]{0.49\linewidth}
        \centering
        \includegraphics[width=\textwidth]{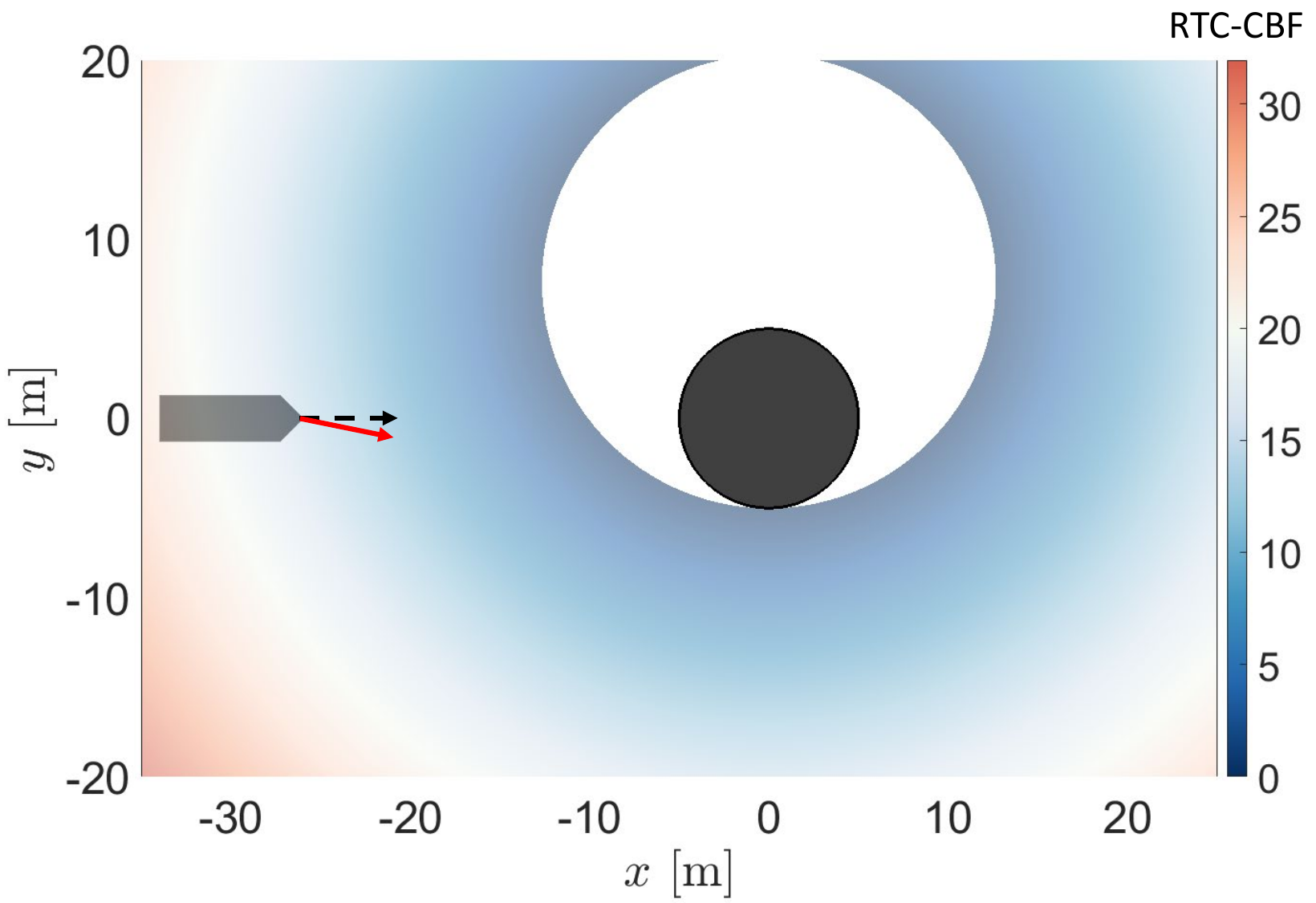}
        \caption{RTC-CBF.}
        \label{fig:rcbf}
    \end{subfigure}
    \begin{subfigure}[h]{0.49\linewidth}
        \centering
        \includegraphics[width=\textwidth]{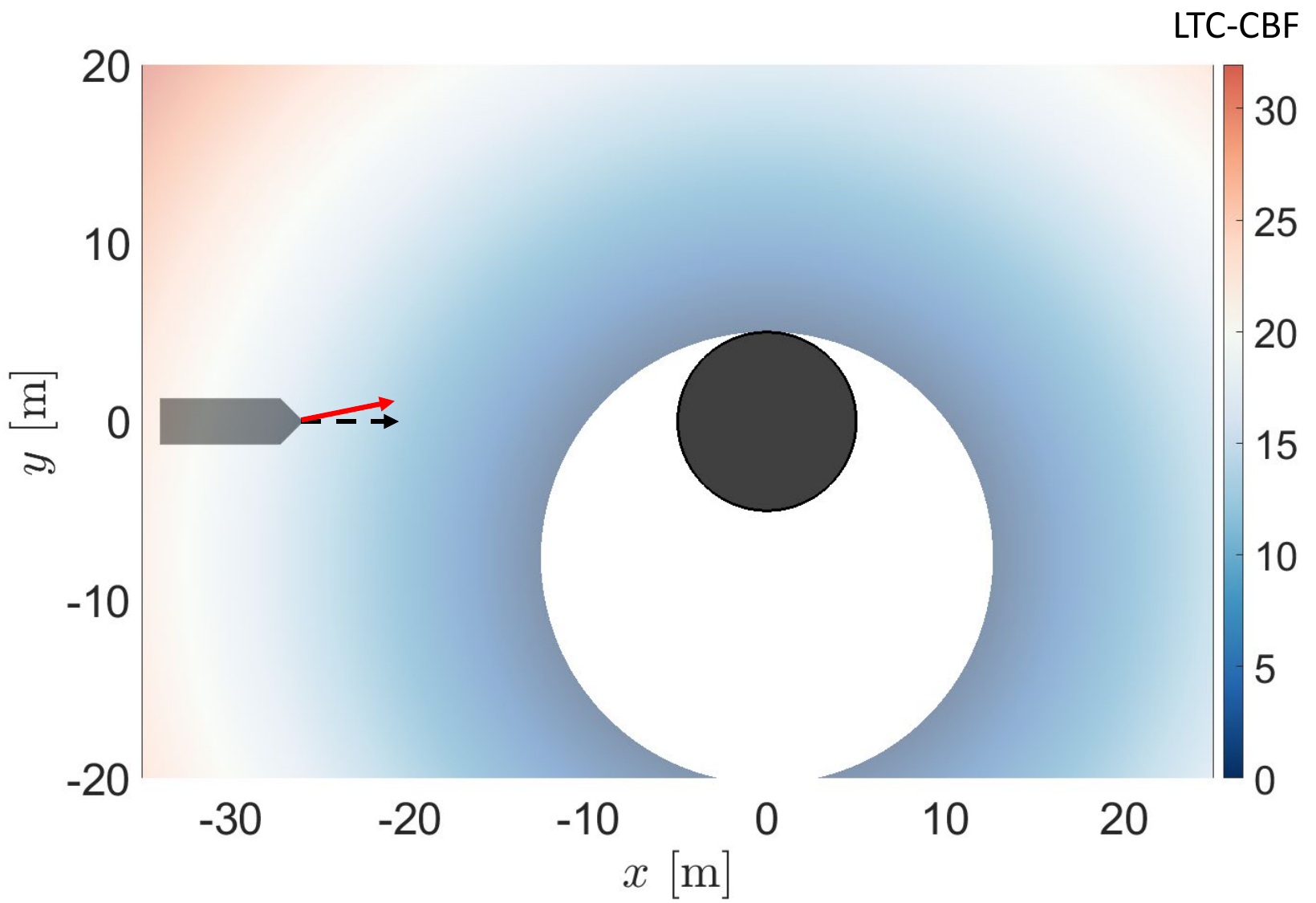}
        \caption{LTC-CBF.}
        \label{fig:lcbf}
    \end{subfigure}    
    \caption{Visualization of CBF constraints. Dashed black arrows represent nominal control inputs, while red arrows indicate safe control inputs filtered by QP with TC-CBF constraints.}
    \label{fig:cbf_visualize}
\end{figure*}

\begin{figure*}[t]   
\centerline{\includegraphics[width=\linewidth]{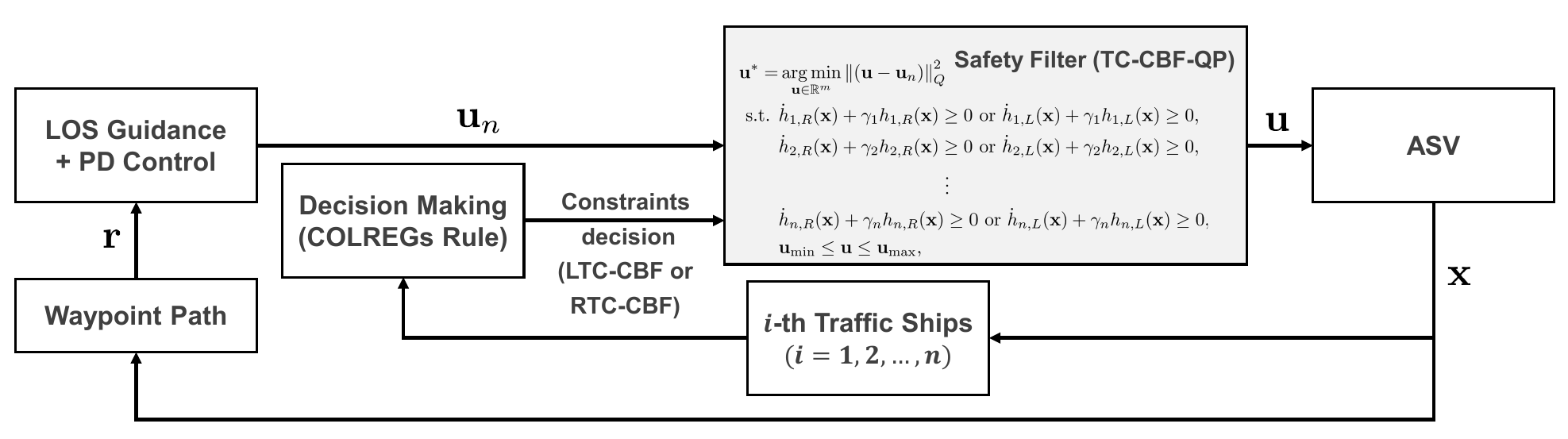}}
    \caption{Proposed TC-CBF-QP-based COLREGs compliant collision avoidance framework.}
    \label{fig:flowchart}
\end{figure*}

\subsection{Turning circle-based CBF design}
The TC-CBF is proposed to address the limitations of existing CBF algorithms that cannot assign an avoidance direction or consider the turning capabilities of ships. The TC-CBF is based on the principle that a vehicle can avoid obstacles as long as its turning circle, determined by its maximum turning capability, current speed, and heading, does not intersect with obstacles.
The TC-CBF for each direction is defined $h_{R}$ for the right direction and $h_{L}$ for the left direction, as follows:
\begin{equation}
\begin{aligned}
    h_{R} (\mathbf{x}) &= d_R^2 - (o_r+R_s+ R)^2, \\
    h_{L} (\mathbf{x}) &= d_L^2 - (o_r+R_s+ R)^2,
    \label{eq:leftright}
\end{aligned}
\end{equation}
where $d_R$ and $d_L$ represent the distances from the centers of the right and left-turning circles to the obstacles, respectively. The centers of the turning circles and obstacles are given as:
\begin{equation}
    \mathbf{p}_{R} = [x_{R}, y_{R}]^\top, \     \mathbf{p}_{L} = [x_{L}, y_{L}]^\top, \     \mathbf{o} = [o_{x}, o_{y}]^\top. \  
\end{equation}
The distances between the turning circles and obstacles can be obtained as follows:
\begin{equation}
d_R = || \mathbf{p}_{R} - \mathbf{o} ||,\
d_L = || \mathbf{p}_{L} - \mathbf{o} ||.
\end{equation}
The radius of the turning circle, $R$, is defined as:
\begin{equation}
    R = \alpha  \frac{u}{r_{\text{max}}},
\end{equation}
where $u$ is the speed of the vehicle, $r_{\text{max}} = \max(r)$ denotes the maximum turning rate, and $\alpha$ is a scaling factor used to smooth the avoidance maneuver by increasing the turning radius. $o_r$ and $R_S$ are the safe radii of the traffic and own ships, respectively. In this study, we assume that two values are identical, i.e., $o_r = R_S$. 
The center positions of the turning circles are given by:
\begin{equation}
\begin{aligned}    
    (x_{R},y_{R})  &= (x + R\cos(\psi-\pi/2), y+R\sin(\psi-\pi/2)), \\
    (x_{L},y_{L})  &= (x + R\cos(\psi+\pi/2), y+R\sin(\psi+\pi/2)).
\end{aligned}
\label{eq:tccbf_form}
\end{equation}
Two turning circles are depicted in Fig.~\ref{fig:turning_circles}.
Non-negative values of $h_{R}$ and $h_{L}$ indicate that the vehicle's right and left turning circles maintain a safe distance from obstacles. As a result, $h_{R}$ ensures safe avoidance on the right side, while $h_{L}$ ensures safe avoidance on the left side, considering the vehicle's turning capabilities. This method provides an effective way to adhere to COLREGs rules.

Figure~\ref{fig:cbf_visualize} visualizes the values of the two TC-CBFs. Obstacles are depicted in black, while the white regions where CBF values are zero indicate areas that must be avoided to maintain the current speed and heading under the constraints of the CBF. 
The size and shape of these restricted regions vary with the vehicle's speed and its orientation relative to the obstacle. As shown in the figure, the LTC-CBF forces the vehicle to avoid the left side, and the RTC-CBF guides it to the right side, respectively.

\subsection{TC-CBF-QP Safety Filter}
To implement a safety filter using the proposed TC-CBFs, the following CBF constraints are introduced:
\begin{equation}
\begin{aligned}
   \dot{h}_{R}(\mathbf{x}) + \gamma{h}_{R}(\mathbf{x}) & \geq 0, \\
   \dot{h}_{L}(\mathbf{x}) + \gamma{h}_{L}(\mathbf{x}) & \geq 0,
\end{aligned}
\end{equation}
where the functions $h_R$ and $h_L$ can be rewritten as follows:
\begin{equation}
\begin{aligned}    
    h_{R} &= || \mathbf{p}_{R} - \mathbf{o} || ^2 - (o_r+ R_s+R)^2, \\
    h_{L} &= || \mathbf{p}_{L} - \mathbf{o} || ^2 - (o_r+ R_s+R)^2.
\end{aligned}
\end{equation}
Differentiating $h_{R}$ and $h_{L}$ with respect to time yields:
\begin{equation}
\begin{aligned}
\dot{h}_{R}(\mathbf{x}) &= 2( \mathbf{p}_{R} - \mathbf{o} )^\top ( \dot{\mathbf{p}}_{R} - \dot{\mathbf{o}} ) - 2(o_r + R_s+R)\dot{R}, \\
\dot{h}_{L}(\mathbf{x}) &= 2( \mathbf{p}_{L} - \mathbf{o} )^\top ( \dot{\mathbf{p}}_{L} - \dot{\mathbf{o}} ) - 2(o_r + R_s+R)\dot{R}.
\end{aligned}
\end{equation}
Based on the encounter situation decision-making module, if the own ship needs to avoid the $i$-th traffic ship on the left side, the LTC-CBF constraints are applied; for all other cases, the RTC-CBF constraints are used. If there are $n$ traffic ships nearby, the following optimization problem can be formulated using the nominal control input $\mathbf{u}_n = [a_n, r_n]^\top$, obtained from equations \eqref{eq:acc_pd} and \eqref{eq:rot_pd}, as follows:
\begin{equation}
\begin{aligned}
 \mathbf{u}^* = & \argmin_{\mathbf{u}\in\mathbb{R}^m} \left\|\left(\mathbf{u}-\mathbf{u}_{n}\right)\right\|^2_Q \\
\text { s.t. } & 
   \dot{h}_{1,R}(\mathbf{x}) + \gamma {h}_{1,R}(\mathbf{x}) \geq 0 \ \text{or} \ 
   \dot{h}_{1,L}(\mathbf{x}) + \gamma {h}_{1,L}(\mathbf{x}) \geq 0, \\ 
&  \dot{h}_{2,R}(\mathbf{x}) + \gamma {h}_{2,R}(\mathbf{x}) \geq 0 \ \text{or} \ 
   \dot{h}_{2,L}(\mathbf{x}) + \gamma {h}_{2,L}(\mathbf{x}) \geq 0, \\ 
   & \qquad \qquad \qquad \quad \quad \quad \vdots \\
   & \dot{h}_{n,R}(\mathbf{x}) + \gamma {h}_{n,R}(\mathbf{x}) \geq 0 \ \text{or} \ 
   \dot{h}_{n,L}(\mathbf{x}) + \gamma {h}_{n,L}(\mathbf{x}) \geq 0, \\ 
   &\mathbf{u}_{\text{min}} \leq \mathbf{u} \leq \mathbf{u}_{\text{max}},
\end{aligned}
\end{equation}
where $Q\in \mathbb{R}^{2\times 2}$ is a weighting matrix, and $h_{i,R}$ and $h_{i,L}$ represent the TC-CBFs for the $i$-th traffic ship, depending on encounter type. If the own ship needs to avoid the $i$-th traffic ship on the left side, the constraints $\dot{h}_{i,L} + \gamma h_{i,L}$ are enforced; for avoidance on the right side, $\dot{h}_{i,R} + \gamma h_{i,R}$ is applied. The last inequality constraint accounts for input saturation. The overall framework of the proposed approach is shown in Fig.~\ref{fig:flowchart}.

\section{Simulation Results} \label{sec:simulation}
To evaluate the computational efficiency and performance of the proposed algorithm, a series of simulation experiments were conducted, divided into three sections.
In the first section, we analyze the impact of two tuning parameters in the proposed CBF under one-to-one encounter situations, focusing on a starboard crossing scenario. The effectiveness of the algorithm is further validated through overtaking and head-on scenarios.
In the second section, we compare the proposed algorithm with an MPC-based algorithm in scenarios involving multiple traffic ship encounters.
Finally, in the third section, we validate the computational times of the proposed algorithm in a highly complex environment, where all traffic ships use the proposed method to avoid each other while adhering to COLREGs.

Our algorithm and the simulations are implemented in a MATLAB environment, and the solver utilizes the alternating direction method of multipliers (ADMM)-based QP solver, OSQP \cite{osqp}. The algorithm runs on an Intel i5-14600KF CPU @ 3.50GHz with 32GB of RAM.

\begin{figure}[t]
    \captionsetup[subfigure]{justification=centering}
        \centering
    \begin{subfigure}[h]{\linewidth}
        \centering
        \includegraphics[width=\textwidth]{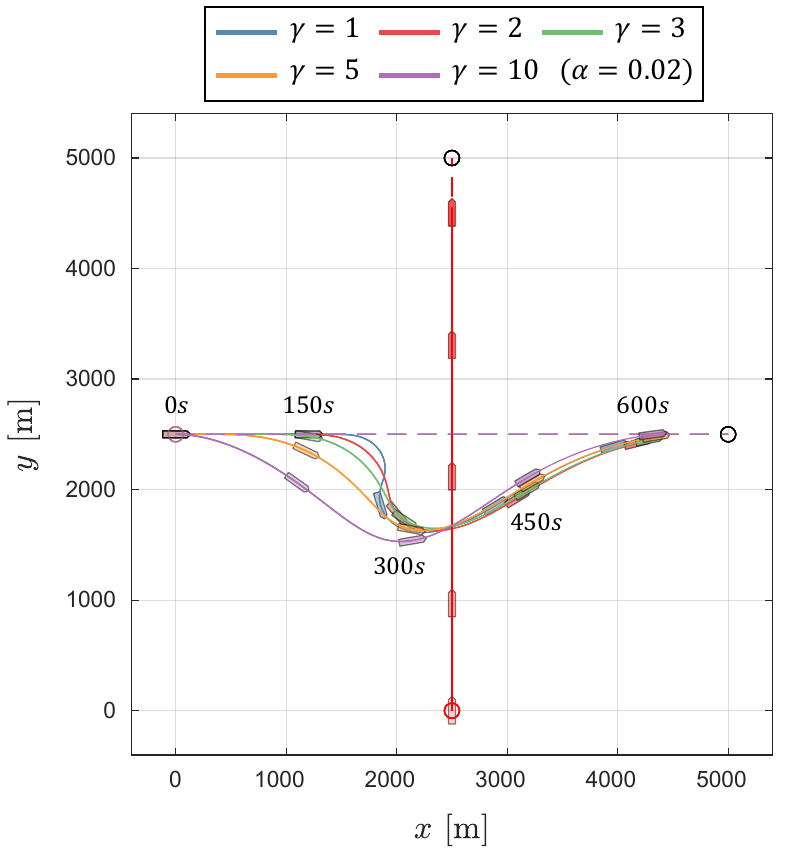}
        \caption{Trajectories of ships.}
        \label{fig:crossing_traj}
    \end{subfigure}
    \begin{subfigure}[h]{\linewidth}
        \centering
        \includegraphics[width=\textwidth]{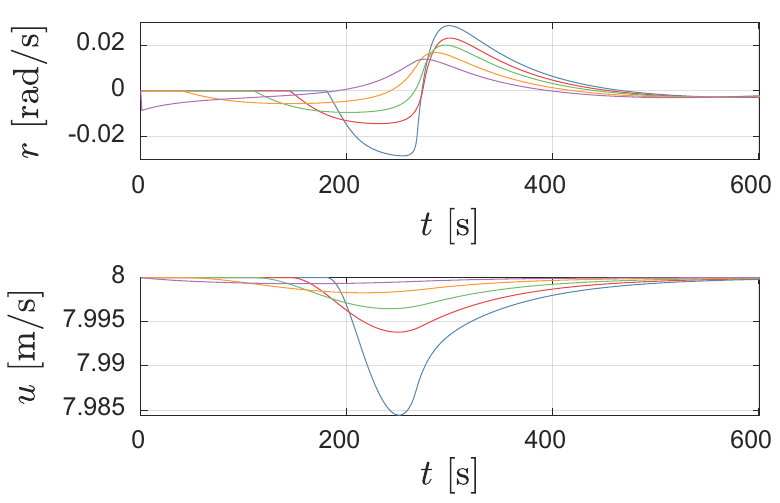}
        \caption{Time trajectories of turning rate and velocities.}
        \label{fig:crossing_result}
    \end{subfigure}       
    \caption{Starboard crossing simulation results with varying $\gamma$ values.}
    \label{fig:crossing}
\end{figure}

\begin{figure}[t]
    \captionsetup[subfigure]{justification=centering}
        \centering
    \begin{subfigure}[h]{\linewidth}
        \centering
        \includegraphics[width=\textwidth]{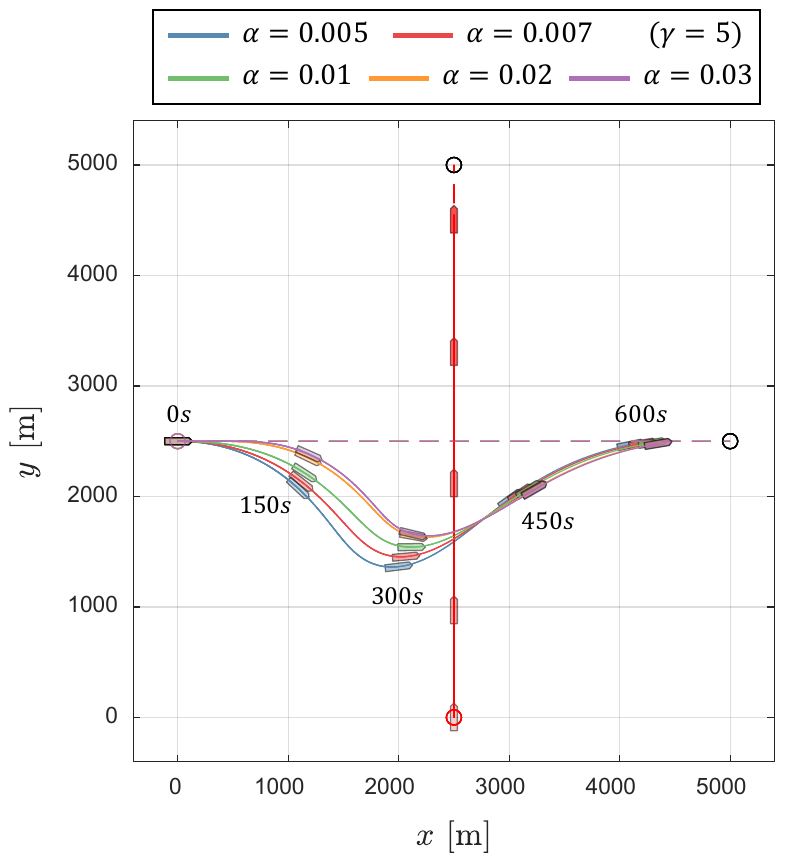}
        \caption{Trajectories of ships.}
        \label{fig:crossing_traj2}
    \end{subfigure}
    \begin{subfigure}[h]{\linewidth}
        \centering
        \includegraphics[width=\textwidth]{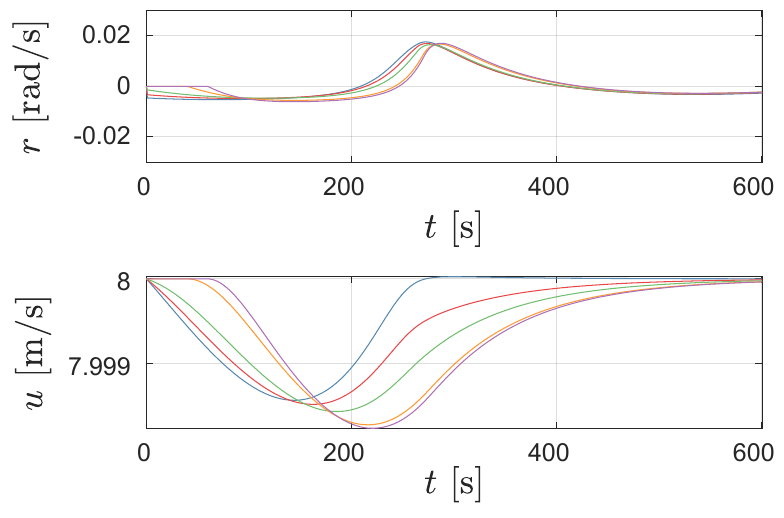}
        \caption{Time trajectories of turning rate and velocities.}
        \label{fig:crossing_result2}
    \end{subfigure}       
    \caption{Starboard crossing simulation results with varying $\alpha$ values.}
    \label{fig:crossing2}
\end{figure}

\subsubsection{One-to-one encounter simulation}
In this section, the control frequency was set to \SI{1}{Hz}, with a sampling time of \SI{1}{\second}, and the desired velocity was set to \SI{8.0}{\meter/\second}. The maximum acceleration and turning rate were limited to \SI{0.03}{\meter/\second^2} and \SI{0.03}{\radian/\second}, respectively. The ship’s length was \SI{25}{\meter}, the LOS distance was \SI{1000}{\meter}, and the safe radius was defined as \SI{250}{\meter}.

We first examine the effect of two key tuning parameters, $\alpha$ and $\gamma$. In the starboard crossing scenario, $\alpha$ was fixed at 0.02, while $\gamma$ was varied across the values $1,\ 2,\ 3,\ 5,\ 10$. The resulting trajectories are shown in Fig.~\ref{fig:crossing_traj}, and the corresponding control results are presented in Fig.~\ref{fig:crossing_result}. As illustrated in Fig.~\ref{fig:crossing}, increasing $\gamma$ leads to earlier avoidance, smoother trajectories, and lower peak turning rates. 
Next, we analyze the impact of $\alpha$ while keeping $\gamma$ fixed at 5. Five values of $\alpha$ were tested: $0.005,\ 0.007,\ 0.01,\ 0.02$, and $0.03$.
The resulting trajectories and control responses are shown in Figs.~\ref{fig:crossing_traj2} and \ref{fig:crossing_result2}, respectively. For a fixed $\gamma$, the ship initiates avoidance at a similar point regardless of $\alpha$, but smaller values of $\alpha$ result in more conservative maneuvers. These results highlight the importance of tuning both $\alpha$ and $\gamma$ in accordance with the desired separation margin, control input constraints, and overall mission objectives.

\begin{figure}[t]
    \captionsetup[subfigure]{justification=centering}
        \centering
    \begin{subfigure}[h]{\linewidth}
        \centering
        \includegraphics[width=\textwidth]{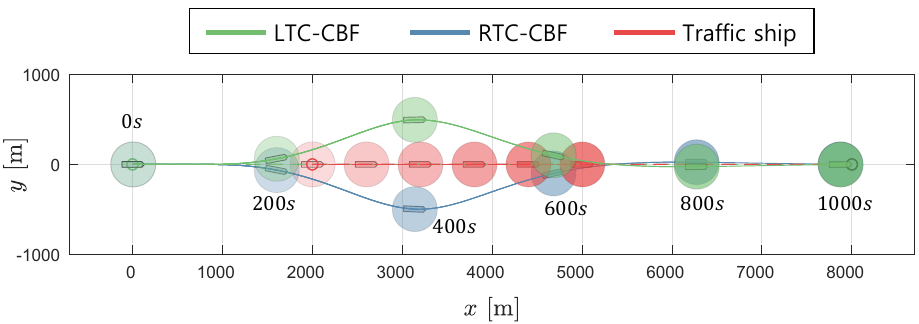}
        \caption{Trajectories of ships.}
        \label{fig:overtaking_traj}
        \vspace{0.3cm}
    \end{subfigure}
    \begin{subfigure}[h]{\linewidth}
        \centering
        \includegraphics[width=\textwidth]{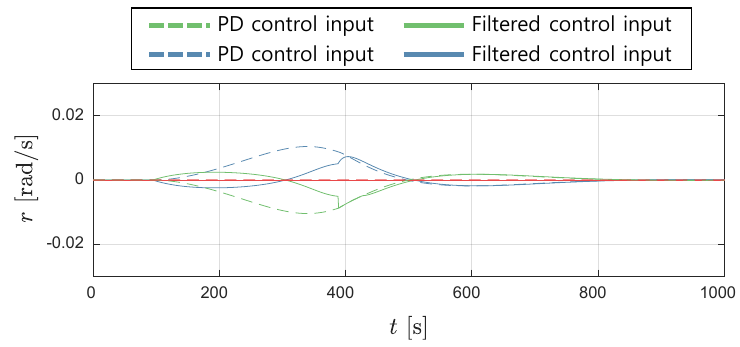}
        \caption{Time trajectories of turning rates.}
        \label{fig:overtaking_input}
    \end{subfigure}       
    \caption{Overtaking scenario simulation.}
    \label{fig:overtaking}
\end{figure}

\begin{figure}[t]
    \captionsetup[subfigure]{justification=centering}
        \centering
    \begin{subfigure}[h]{\linewidth}
        \centering
        \includegraphics[width=\textwidth]{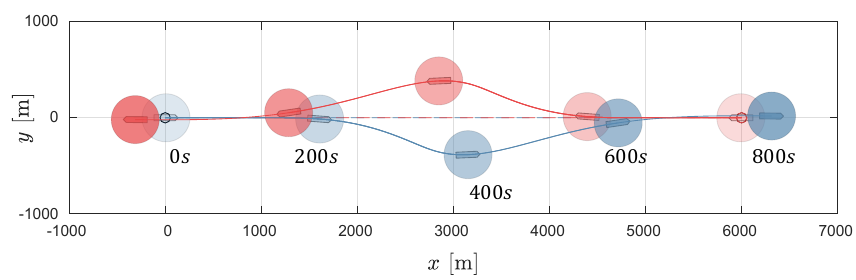}
        \caption{Trajectories of ships.}
        \label{fig:headon_traj}
        \vspace{0.3cm}
    \end{subfigure}
    \begin{subfigure}[h]{\linewidth}
        \centering
        \includegraphics[width=\textwidth]{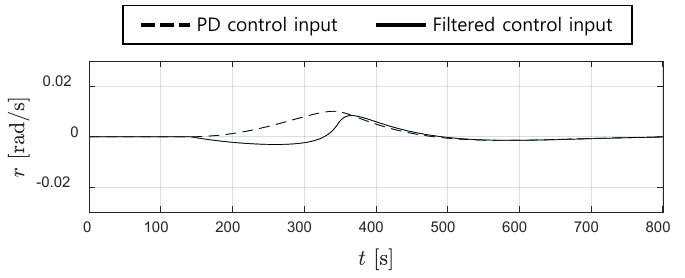}
        \caption{Time trajectories of turning rates.}
        \label{fig:headon_input}
    \end{subfigure}       
    \caption{Head-on scenario simulation.}
    \label{fig:headon}
\end{figure}

To further validate the proposed approach, two additional scenarios were simulated: overtaking and head-on encounters. The results of the overtaking scenario are shown in Fig.~\ref{fig:overtaking_traj}, where the green trajectory corresponds to the LTC-CBF algorithm, and the blue trajectory corresponds to the RTC-CBF algorithm, guiding the own ship to avoid traffic ships on the left and right sides, respectively. The control input trajectories are shown in Fig.~\ref{fig:overtaking_input}. In this figure, the dashed lines represent the control input from the waypoint PD tracking controller defined in \eqref{eq:acc_pd} and \eqref{eq:rot_pd}, while the solid lines represent the final inputs filtered through the proposed TC-CBF-QP algorithm. 
In the head-on scenario, both traffic ships use the proposed algorithm to avoid each other, as shown in Fig.~\ref{fig:headon}. Their trajectories are depicted in Fig.~\ref{fig:headon_traj}, and the corresponding inputs are shown in Fig.~\ref{fig:headon_input}. Since the same CBF parameters are applied to both ships, they follow identical avoidance trajectories and control inputs. If different parameters are used, the ship with a smaller $\alpha$ and larger $\gamma$ will initiate avoidance earlier. These parameters can be customized to account for differences in ship size and operational context.

\begin{figure*}[h]
    \captionsetup[subfigure]{justification=centering}
        \centering
    \begin{subfigure}[h]{0.49\linewidth}
        \centering
        \includegraphics[width=\textwidth]{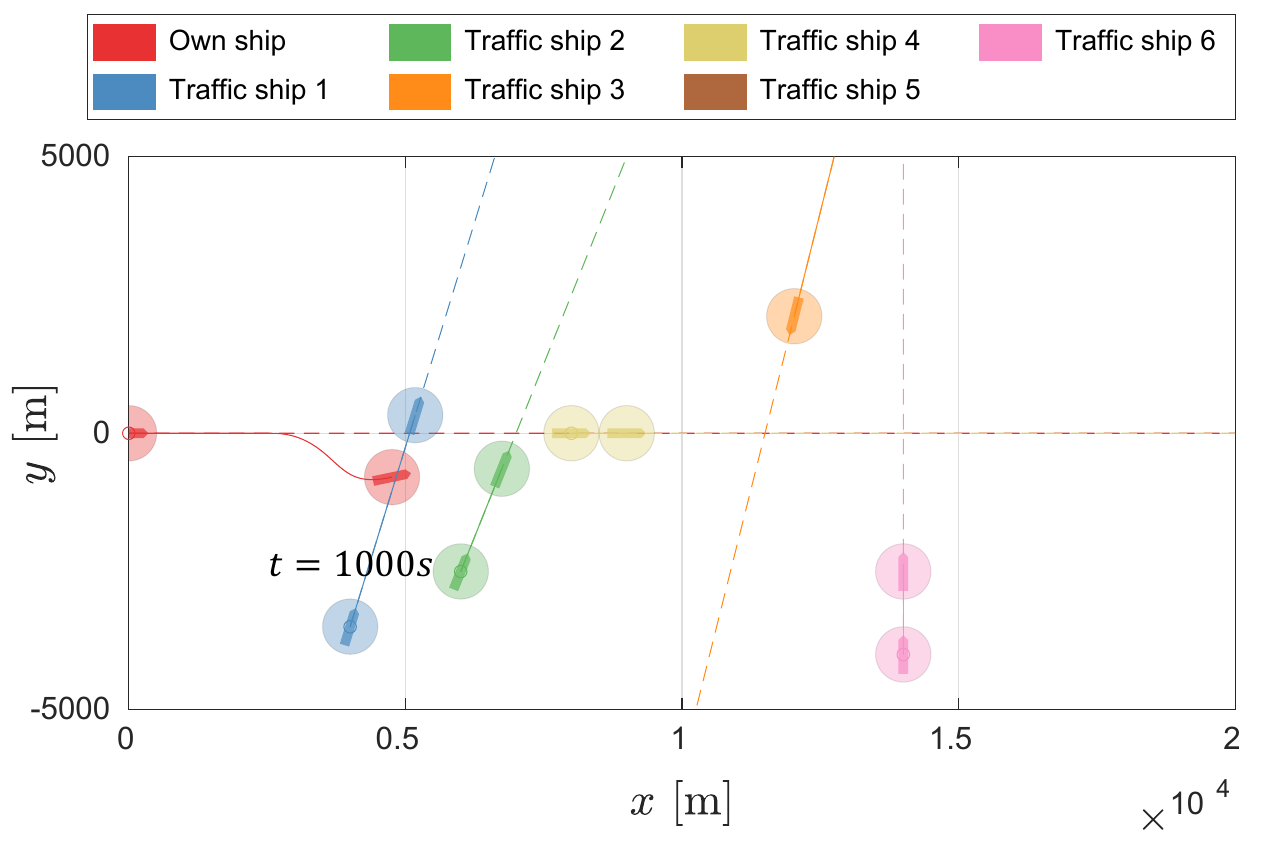}
        \caption{$t=\SI{1000}{\second}$.}
    \end{subfigure}
    \begin{subfigure}[h]{0.49\linewidth}
        \centering
        \includegraphics[width=\textwidth]{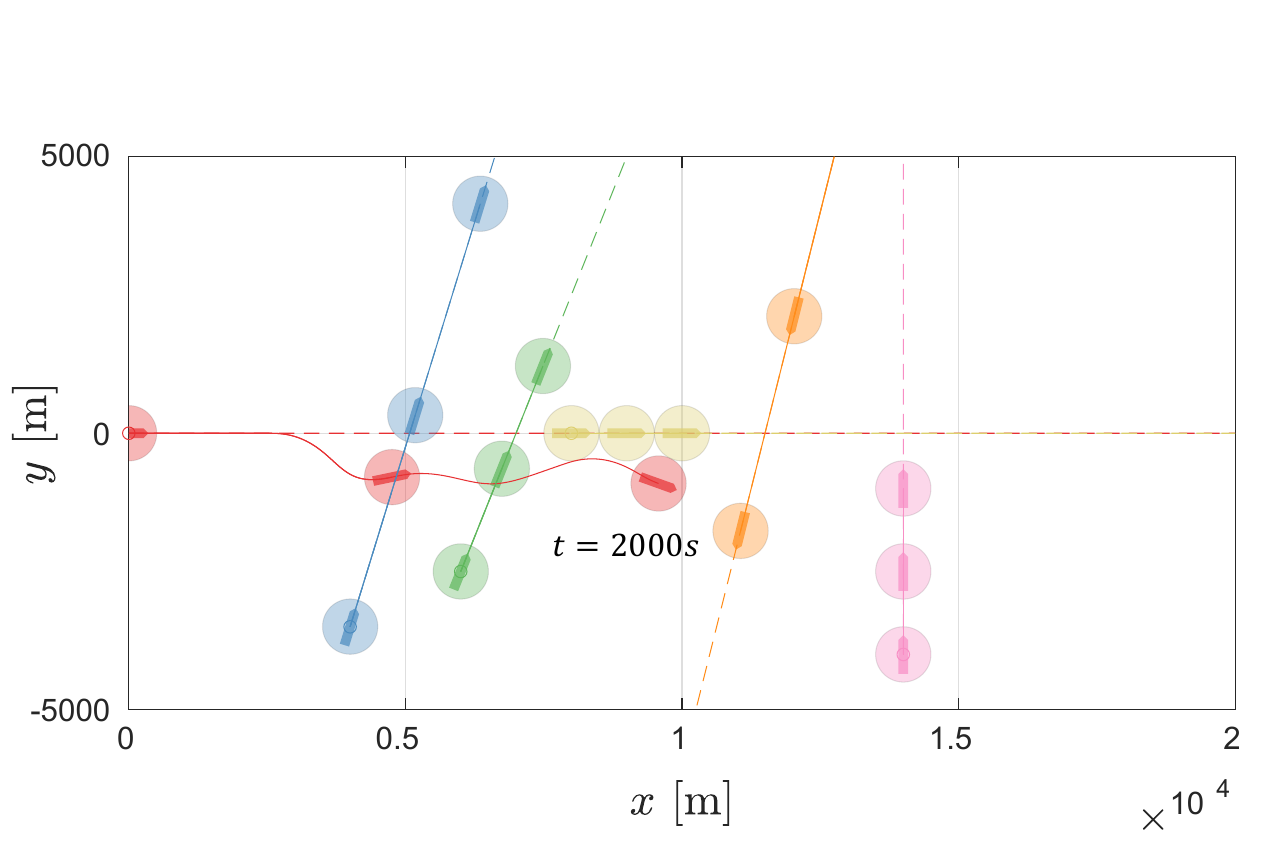}
        \caption{$t=\SI{2000}{\second}$.}
    \end{subfigure}       
    \begin{subfigure}[h]{0.49\linewidth}
        \centering
        \includegraphics[width=\textwidth]{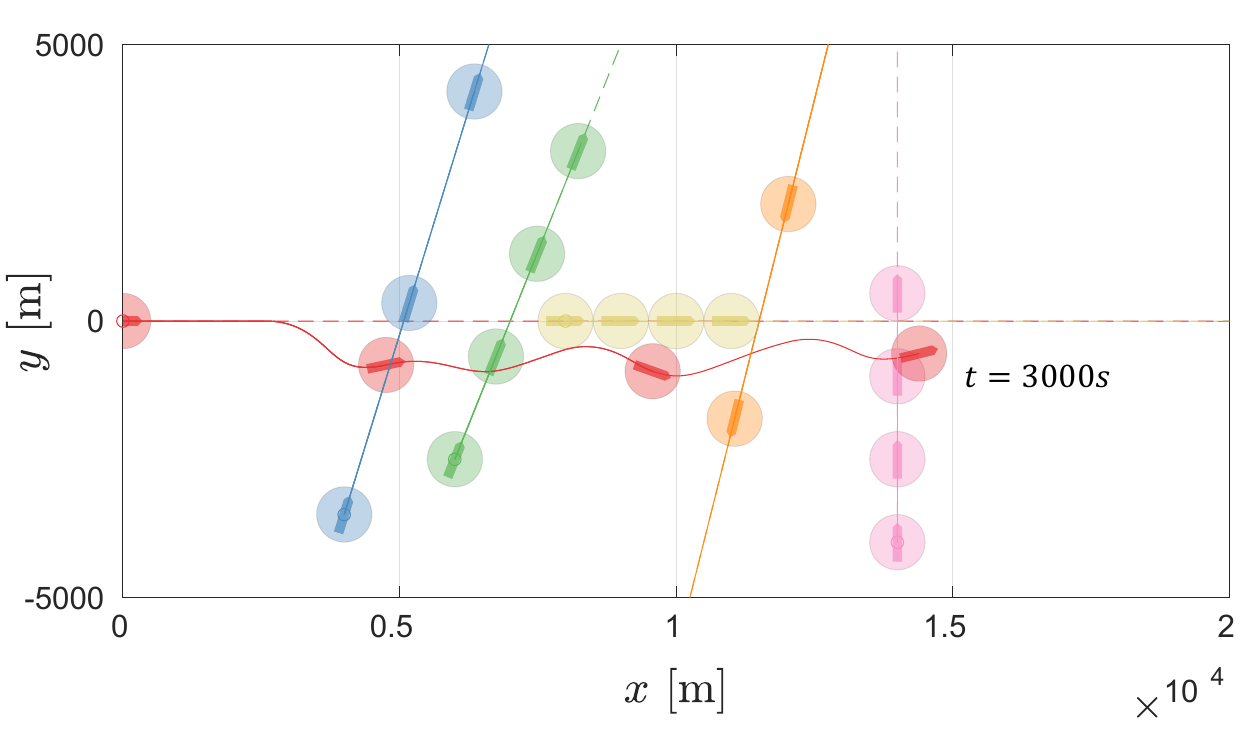}
        \caption{$t=\SI{3000}{\second}$.}
   \end{subfigure}       
    \begin{subfigure}[h]{0.49\linewidth}
        \centering
        \includegraphics[width=\textwidth]{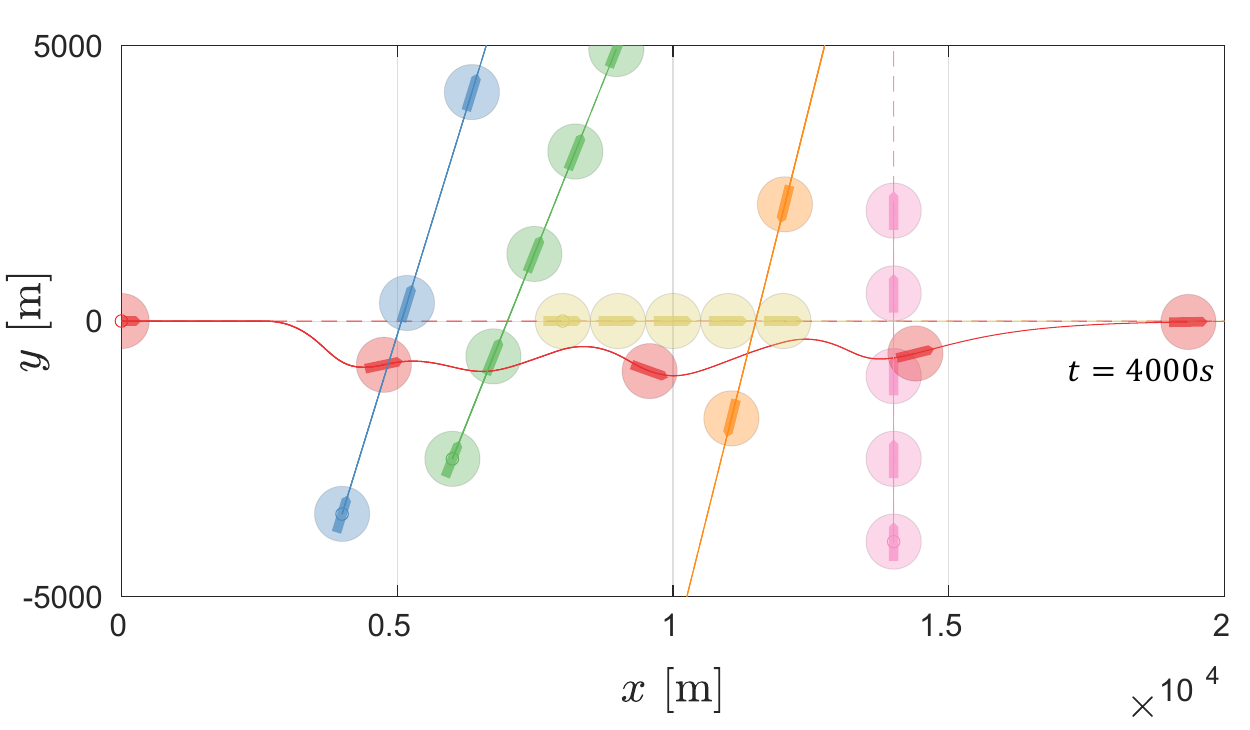}
        \caption{$t=\SI{4000}{\second}$.}
   \end{subfigure}       
    \caption{Multi-ship encounter simulation using the proposed algorithm. (A video demonstration is available at \url{https://youtu.be/HvF8yXl9-hc}.)}     
    \label{fig:multi_encounter_cbf}
\end{figure*}

\begin{figure}[t!]
    \centering
    \includegraphics[width=\linewidth]{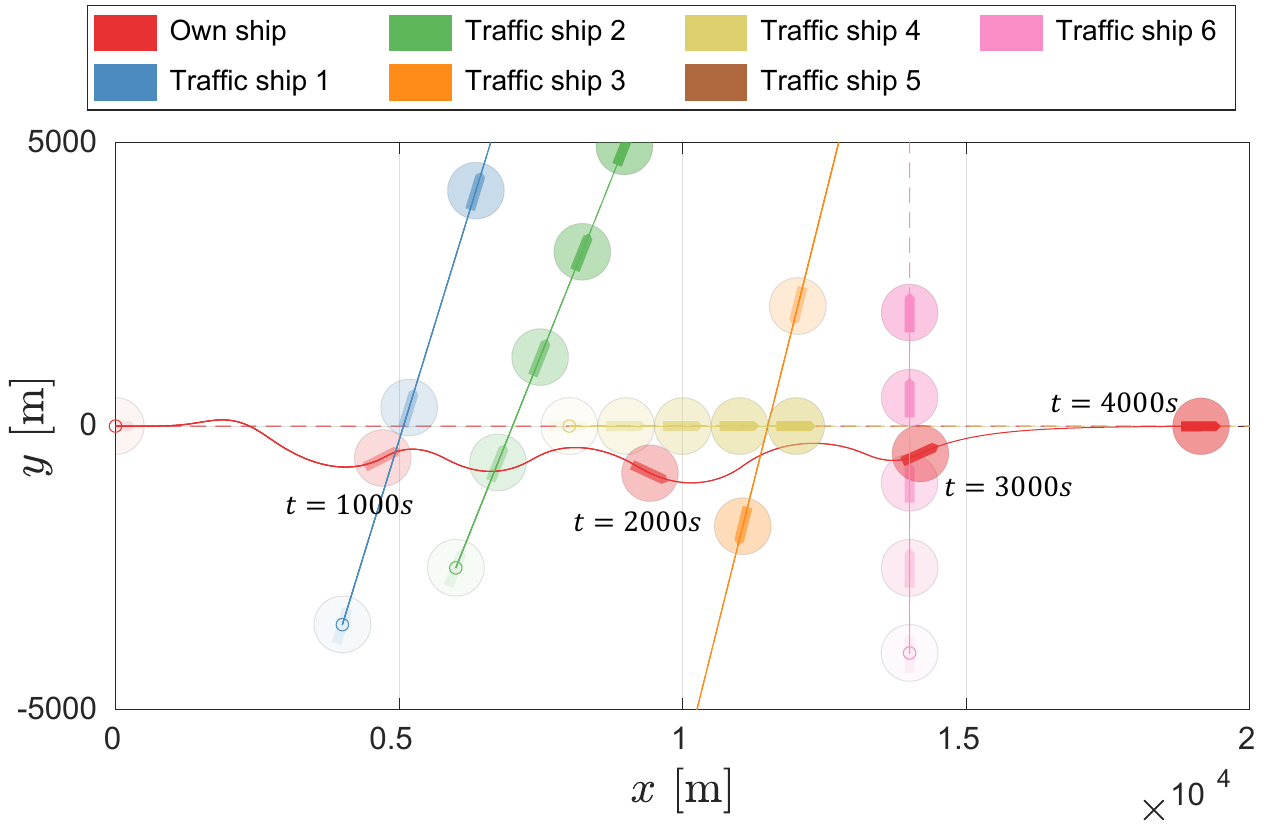}
    \caption{Multi-traffic ship encounter simulation using the MPC-based algorithm.}
    \label{fig:multi_encounter_mpc}
\end{figure}

\begin{figure}[t!]
    \captionsetup[subfigure]{justification=centering}
        \centering
    \begin{subfigure}[h]{0.86\linewidth}
        \centering
        \includegraphics[width=\textwidth]{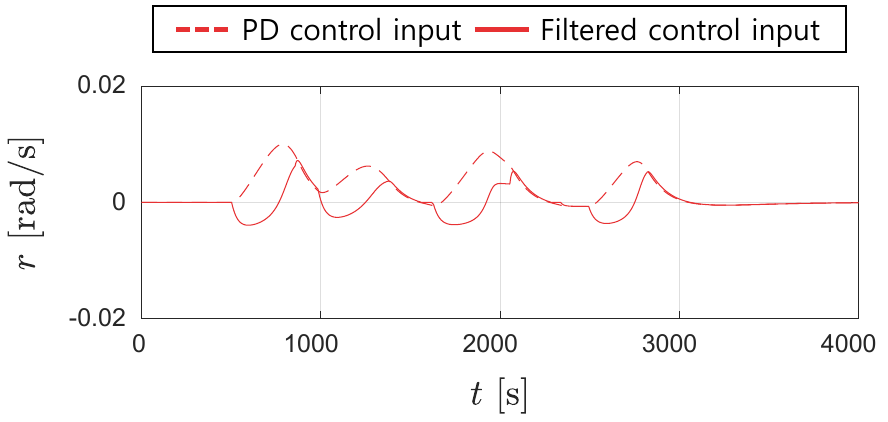}
        \caption{Proposed algorithm.}
        \vspace{0.125cm}
        \label{fig:multi_encounter_cbf_input}
    \end{subfigure}
    \begin{subfigure}[h]{0.86\linewidth}
        \centering
        \includegraphics[width=\textwidth]{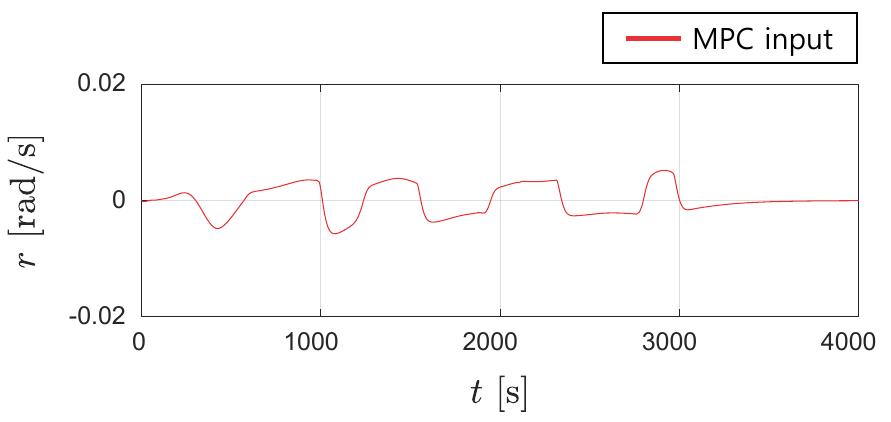}
        \caption{MPC algorithm.}
        \label{fig:multi_encounter_mpc_input}
    \end{subfigure}       
    \caption{Control input trajectories in multi-traffic ships encounter simulation.}
    \label{fig:multi_encounter_input}
\end{figure}

\begin{figure*}[t]
    \captionsetup[subfigure]{justification=centering}
        \centering
    \begin{subfigure}[h]{0.49\linewidth}
        \centering
        \includegraphics[width=0.95\textwidth]{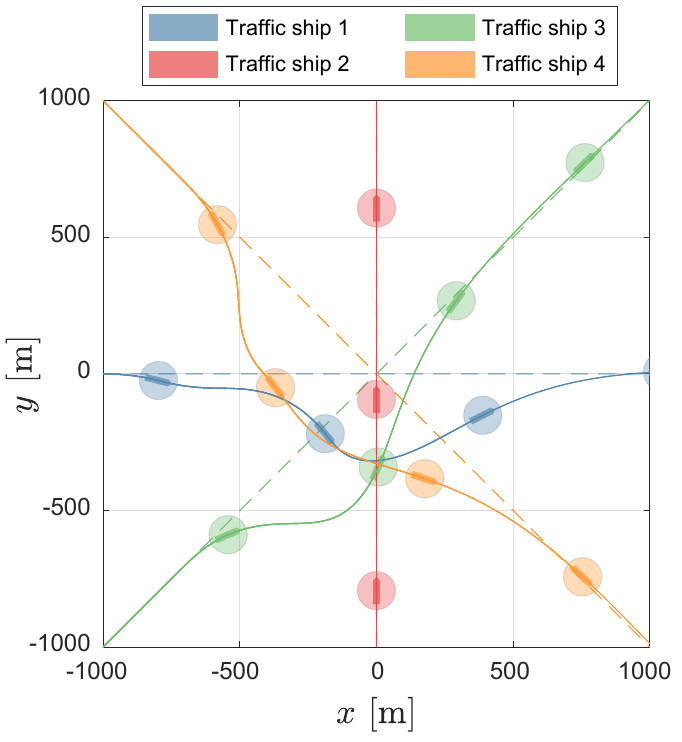}
        \caption{Proposed algorithm.}
        \label{fig:one_point_cbf}
    \end{subfigure}       
    \begin{subfigure}[h]{0.49\linewidth}
        \centering
        \includegraphics[width=0.95\textwidth]{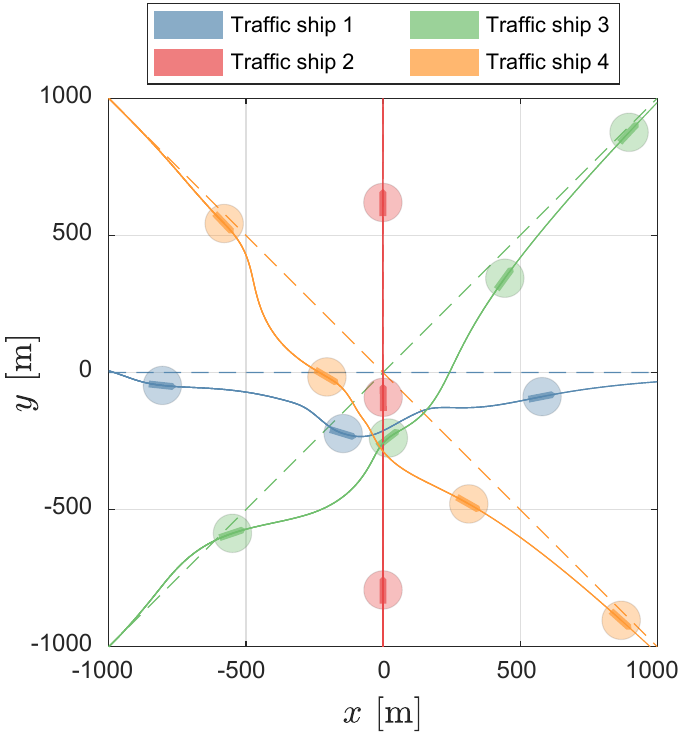}
        \caption{MPC algorithm.}
        \label{fig:one_point_mpc}
    \end{subfigure}
    \caption{Simulation results of the circular formation scenario. Snapshots are taken at \SI{100}{\second} intervals. }
    \label{fig:one_point_result}
\end{figure*}

\subsubsection{Multi-traffic ships encounter simulation}
In this section, we compare the performance of the proposed approach with a state-of-the-art MPC-based algorithm using two scenarios. 
In the first scenario, the simulation setup assumes that the own ship operates at a desired velocity of \SI{5.0}{\meter/\second}, with a maximum acceleration and turning rate of \SI{0.05}{\meter/\second^2} and \SI{0.05}{\radian/\second}, respectively. The own ship's length is \SI{50}{\meter}, the LOS distance is \SI{2000}{\meter}, and the safe radius is \SI{500}{\meter}. We set $\alpha = 10$ and $\gamma = 0.03$, with thresholds for DCPA, TCPA, and distance set to \SI{1500}{\meter}, \SI{1000}{\second}, and \SI{5000}{\meter}, respectively. 
The six traffic ships maintain constant velocity and heading throughout the simulation, while the own ship avoids them. The baseline algorithm, an MPC-based approach, uses ellipse-shaped obstacle avoidance constraints to guide the ship’s avoidance maneuvers according to COLREGs rules. Details of the MPC algorithm are provided in \hyperref[sec:appendix]{Appendix}. The MPC setup includes a prediction horizon of 40 and a sampling time of \SI{15}{\second}, chosen to consider the speed and size of the ships.

Figure~\ref{fig:multi_encounter_cbf} presents the results of the proposed algorithm, while the corresponding control inputs are shown in Fig.~\ref{fig:multi_encounter_cbf_input}. The dashed lines represent the PD control inputs, and the solid red lines indicate the filtered inputs generated by the proposed TC-CBF-QP algorithm. Figure~\ref{fig:multi_encounter_mpc} illustrates the trajectories produced by the MPC-based method, with Fig.~\ref{fig:multi_encounter_mpc_input} displaying the corresponding control inputs. Overall, both algorithms yield similar trajectory and control input profiles.
For a quantitative comparison, we evaluated the average and maximum computational time, the average and maximum cross-track error (CTE), and the total control effort, which was computed as the cumulative sum of the absolute values of accelerations and turning rates. As summarized in Table~\ref{table:computation_time_multi_encounter}, while the MPC-based algorithm achieves higher tracking accuracy, it incurs a greater control effort. 
Despite the superior tracking performance of the MPC-based method, the overall performance of both approaches is comparable when accounting for control costs. As is widely recognized, there exists a trade-off between tracking accuracy and control input usage, which is strongly influenced by parameter tuning. Notably, the proposed method provides a key advantage in terms of computational efficiency, making it particularly effective for real-time applications.

\begin{table*}[t]
\centering
\small
\renewcommand{\arraystretch}{1.43}
    \begin{tabular}{ | c | c | c  | c | c | c | c | }
    \hline
    \textbf{Scenario} &
    \textbf{Controller} & 
    \textbf{Avg. comp. time} & 
    \textbf{Max. comp. time} & 
    \textbf{Avg. CTE} & 
    \textbf{Max. CTE} & 
    \textbf{Control efforts}
    \\     \hline\hline
    \multirow{2}{*}{\makecell{Scenario 1}}  
    & MPC
    & 23.35 [ms]
    & 40.18 [ms]
    & \textbf{390.6} [m]
    & 1001.2 [m]
    & 8.66
    \\ \cdashline{2-7}  %%%%%%%%%%%%%%%%%%%%%%%%%%%%
    & CBF
    & \textbf{0.14} [ms]
    & \textbf{1.80} [ms]
    & 439.5 [m]
    & \textbf{992.4} [m]
    & \textbf{6.16}
    \\ \hline %%%%%%%%%%%%%%%%%%%%%%%%%%%%%%%%%%  
    \multirow{2}{*}{\makecell{Scenario 2}}  
    & MPC
    & 12.60 [ms]
    & 38.70 [ms]
    & \textbf{52.9} [m]
    & \textbf{164.9} [m]
    & 15.29
    \\ \cdashline{2-7}  %%%%%%%%%%%%%%%%%%%%%%%%%%%%
    & CBF
    & \textbf{0.09} [ms]
    & \textbf{1.62} [ms]
    & 61.4 [m]
    & 228.6 [m]
    & \textbf{12.25}
    \\ \hline %%%%%%%%%%%%%%%%%%%%%%%%%%%%%%%%%%      
\end{tabular}
\caption{Performance comparison in multi-ship encounter scenarios.}
\label{table:computation_time_multi_encounter}
\end{table*}

\begin{figure*}[t]
    \captionsetup[subfigure]{justification=centering}
        \centering
    \begin{subfigure}[h]{0.49\linewidth}
        \centering
        \includegraphics[width=\textwidth]{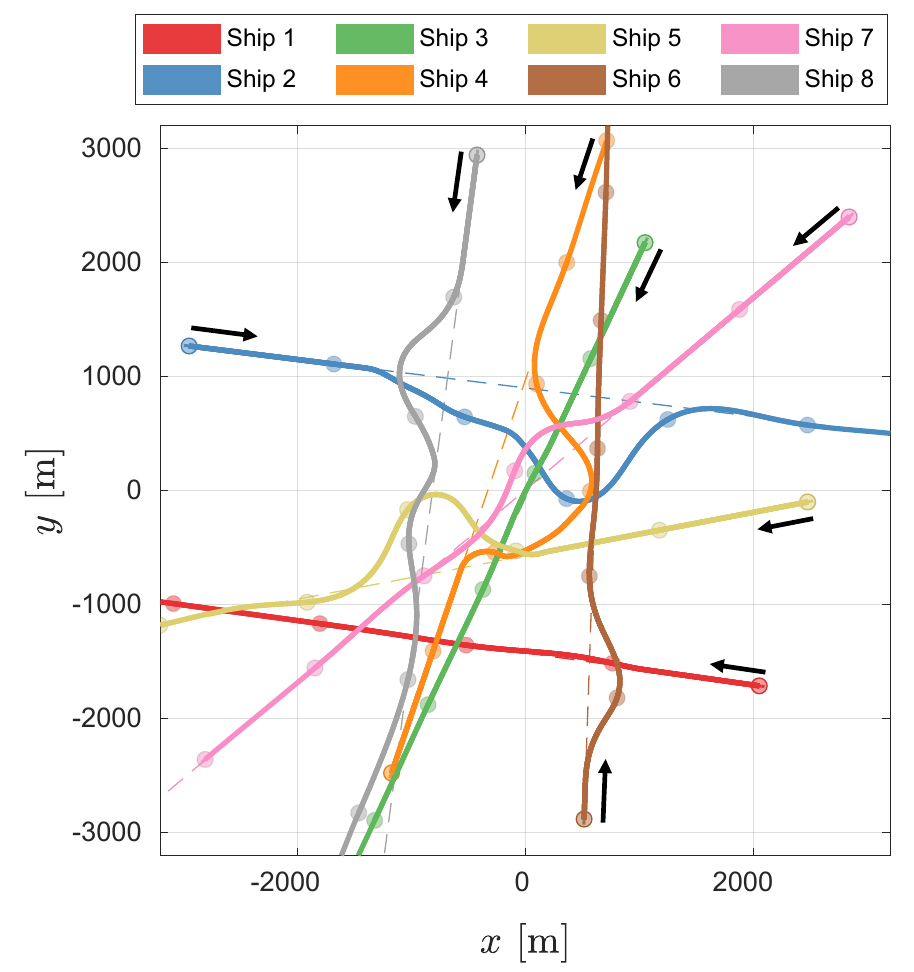}
        \caption{Trajectories from the 8 traffic ship simulation. \\(Additional results available at \url{https://youtu.be/TujWKvx-GDQ})}
        \label{fig:eight_sim}
    \end{subfigure}       
    \begin{subfigure}[h]{0.49\linewidth}
        \centering
        \includegraphics[width=\textwidth]{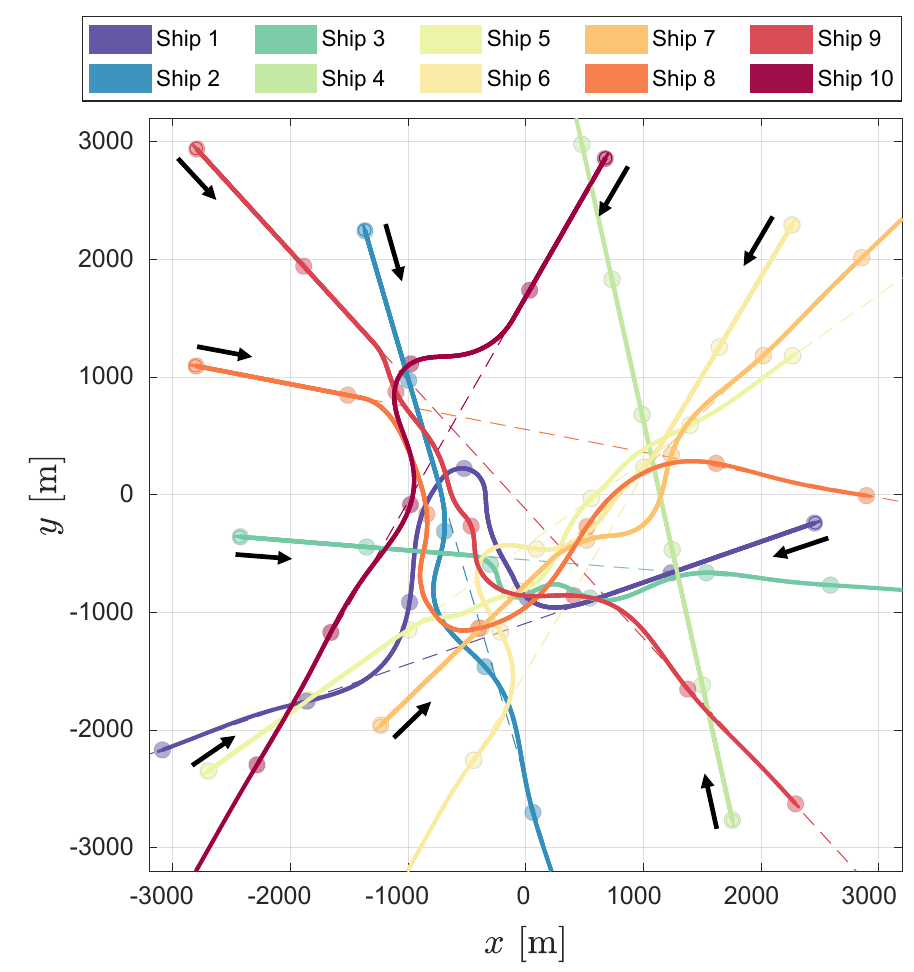}
        \caption{Trajectories from the 10 traffic ship simulation. \\(Additional results available at \url{https://youtu.be/bBjPDJtwvZQ})}
        \label{fig:ten_sim}
    \end{subfigure}
    \caption{Snapshots from simulations in a complex traffic environment. Snapshots are taken at \SI{200}{\second} intervals. Black arrows indicate initial positions and headings.}
    \label{fig:8_10_sims}
\end{figure*}

In the second scenario, four traffic ships utilize both the proposed and the MPC-based algorithms to avoid each other. 
The own ship moves at a desired velocity of \SI{7.0}{\meter/\second}, with the maximum acceleration and turning rate set to \SI{0.05}{\meter/\second^2} and \SI{0.05}{\radian/\second}, respectively. The own ship's length is \SI{7}{\meter}, while the LOS distance and safe radius are set to \SI{400}{\meter} and \SI{70}{\meter}, respectively. The tuning parameters are set to $\alpha = 2$ and $\gamma = 0.05$. For the MPC algorithm, a prediction horizon of 40 steps and a sampling time of \SI{5}{\second} are used.
Thresholds for DCPA, TCPA, and distance are set to \SI{200}{\meter}, \SI{200}{\second}, and \SI{500}{\meter}, respectively.
All four ships begin in a circular formation and move toward the origin. The resulting trajectories are shown in Fig.~\ref{fig:one_point_result}, and a summary of the key performance metrics for both algorithms is provided in Table \ref{table:computation_time_multi_encounter}. As in the first scenario, the proposed algorithm achieves comparable performance while requiring significantly less computational effort.

\subsection{Complex traffic environment simulations}
Finally, we evaluate the efficiency of the proposed approach in highly complex scenarios involving simultaneous collision avoidance with 6, 8, and 10 traffic ships.
In these simulations, each traffic ship’s desired velocity is randomly selected between \SI{5.0}{\meter/\second} and \SI{7.0}{\meter/\second}. The maximum acceleration and turning rate are set to \SI{0.05}{\meter/\second^2} and \SI{0.05}{\radian/\second}, respectively. The own ship’s length is \SI{7}{\meter}, with a LOS distance of \SI{400}{\meter} and a safe radius of \SI{70}{\meter}. The tuning parameters are set to $\alpha =5$ and $\gamma = 0.02$.  Thresholds for DCPA, TCPA, and distance are set to \SI{400}{\meter}, \SI{200}{\second}, and \SI{300}{\meter}, respectively.

Figure~\ref{fig:8_10_sims} illustrates example outcomes from 100 Monte Carlo simulations with 8 and 10 traffic ships, and Table~\ref{table:8_10_sims} summarizes the computational times across all trials. Since the proposed method is formulated as a QP problem, its computational cost remains relatively unchanged even as the number of ships increases. As a result, the algorithm maintains efficient performance and low computational overhead, even under extreme multi-ship encounter conditions.

\begin{table}[h]
\centering
\small
\renewcommand{\arraystretch}{1.43}
    \begin{tabular}{| c | c  | c | }
    \hline
    \textbf{Number of ships} & 
    \textbf{Avg. comp. time} & 
    \textbf{Max. comp. time} 
    \\     \hline\hline
    6
    & 0.111 [ms]
    & 1.934 [ms]
    \\ \hline %%%%%%%%%%%%%%%%%%%%%%%%%%%%%  
    8
    & 0.117 [ms]
    & 1.947 [ms]
    \\ \hline %%%%%%%%%%%%%%%%%%%%%%%%%%%%%
    10
    & 0.128 [ms]
    & 1.962 [ms]
    \\ \hline %%%%%%%%%%%%%%%%%%%%%%%%%%%%%  
\end{tabular}
\caption{Comparison of computation times [ms] in complex traffic environment.}
\label{table:8_10_sims}
\end{table}

\section{Conclusion} \label{sec:conclusion}
This paper presented the TC-CBF-QP-based safety filter algorithm for efficient COLREGs-compliant collision avoidance in autonomous ships. Two types of CBFs, LTC-CBF and RTC-CBF, were introduced to account for the left and right-turning capabilities, respectively. An encounter-type decision-making algorithm selects the appropriate CBF for each identified encounter ship, enabling direction-aware and rule-compliant avoidance behavior.
The proposed algorithm is designed to be compatible with any controller and demonstrates significantly lower computational cost compared to conventional trajectory optimization-based COLREGs-compliant methods. Simulation results across various scenarios demonstrate that the proposed method achieves collision avoidance performance comparable to the MPC-based approach while providing substantial improvements in computational efficiency.

\phantomsection
\addcontentsline{toc}{section}{Appendix}

\section*{Appendix} \label{sec:appendix}
\subsection{MPC-based COLREGs compliant collision avoidance algorithm}
\begin{figure}[t]   
\centerline{\includegraphics[width=\linewidth]{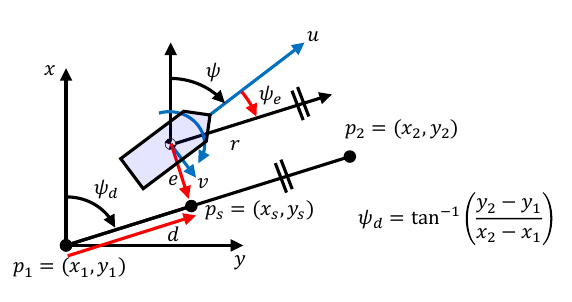}}
    \caption{Path-based coordinate system used in the MPC algorithm.}
    \label{fig:mpc_coord}
\end{figure}

To compare our approach with a state-of-the-art algorithm, we developed an MPC-based COLREGs-compliant collision avoidance algorithm. The formulation is based on the unicycle model, with the state and input vectors defined in the waypoint path coordinate frame as $\mathbf{x} = [d,e,\psi_e,u]^\top$, $\mathbf{u} = [a,r]^\top$, where $d$ is the along-path distance, $e$ is the cross-track error, $\psi_e$ is the heading error, and $u$ is the forward speed. The coordinate system is illustrated in Fig.~\ref{fig:mpc_coord}.
To track the waypoint path while minimizing control effort and respecting the system dynamics, we formulate the following discrete-time optimal control problem:
\begin{subequations}
\label{NMPC} %
\begin{equation}
\min_{\mathbf{x}(\cdot),{\mathbf{u}}(\cdot)}\sum_{i=0}^{N-1}
\ell (\mathbf{x}_i, \mathbf{r}_i, \mathbf{u}_i) + \ell_T(\mathbf{x}_{N},\mathbf{r}_{N})
\end{equation}
\begin{align}
    \text{s.t. } \mathbf{x}_0 - \mathbf{x}_{\text{init}} &= 0,  \label{mpc:init} \\[5pt]
    \mathbf{x}_{i+1} - f_d(\mathbf{x}_i,\mathbf{u}_i) &= 0 , \ i = 0,\ldots,N-1, \label{mpc:dynamics} \\[5pt]
    g(\mathbf{u}_i) & \leq 0, \ i=0,\ldots,N-1, \label{mpc:constraints1} \\[5pt]
    h(\mathbf{x}_i, L_j) & \leq 0, \ i=0,\ldots,N, \ j=0,\ldots,N_{\text{ts}}, \label{mpc:constraints2} 
\end{align}
\end{subequations}
where $ \mathbf{r}_i = [0,0,0,u_d]$ is the reference state, $\mathbf{x}_{\text{init}}$ is the initial state, $f_d$ denotes the vehicle dynamics in the discrete-time domain, $\ell$ and $\ell_T$ are the stage and terminal cost functions, and $N$ is the prediction horizon. The functions \eqref{mpc:constraints1} and \eqref{mpc:constraints2} represent the control input and collision avoidance constraints, respectively. $N_{\text{ts}}$ is the number of traffic ships.

\begin{figure}[t]   
\centerline{\includegraphics[width=\linewidth]{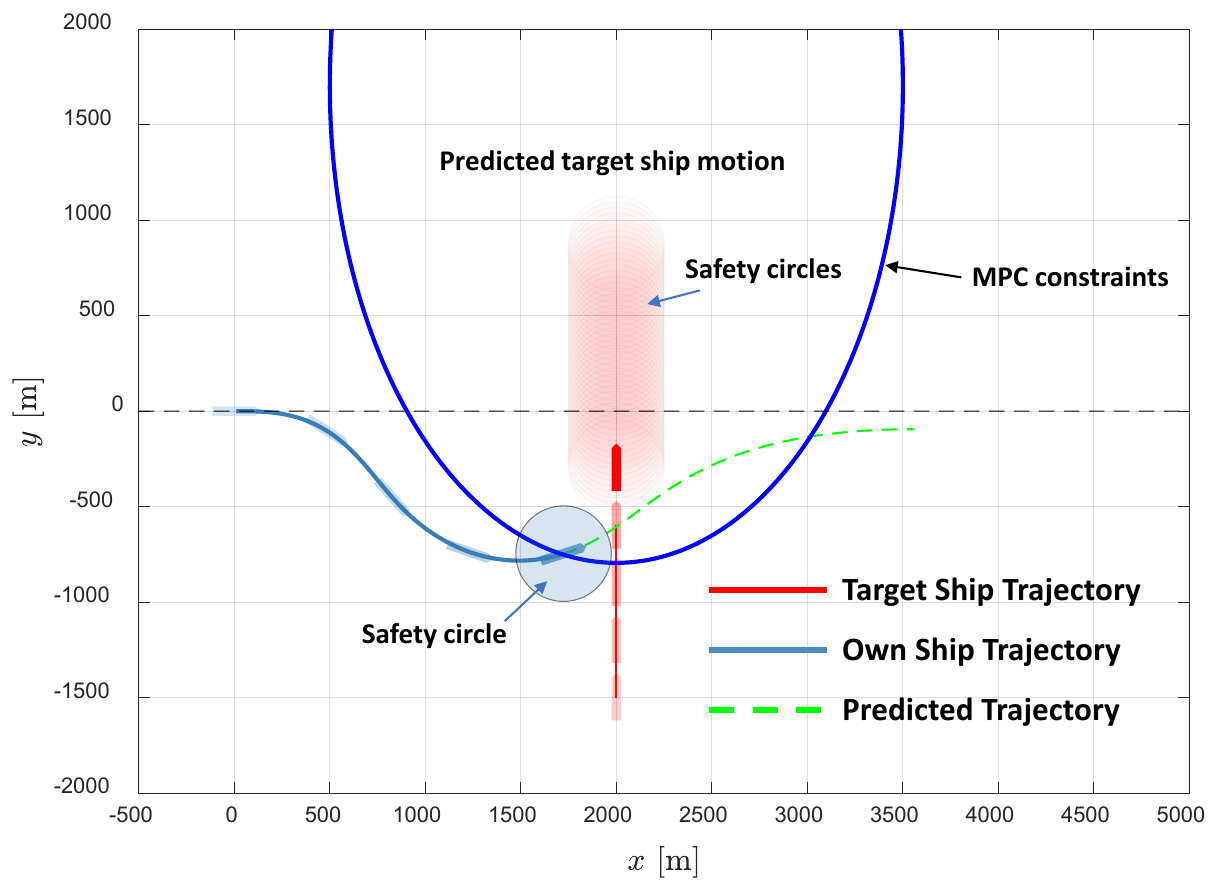}}
    \caption{Visualization of collision avoidance constraints in the MPC algorithm.}
    \label{fig:mpc_ex}
\end{figure}

The cost functions penalize tracking errors and control effort as follows:
\begin{subequations}
\begin{align}
\ell(\mathbf{x}_i, \mathbf{r}_i, \mathbf{u}_i) &= ||\mathbf{x}_i - \mathbf{r}_i||^{2}_Q + ||\mathbf{u}_i||^2_R   + ||\Delta\mathbf{u}_i||^2_{R_d}  \\
\ell_T(\mathbf{x}_{N},\mathbf{r}_{N}) &= ||\mathbf{x}_{N_p} - \mathbf{r}_{N_p}||^2_{Q_T}
\end{align}
\end{subequations}
where $Q$, $R$, $R_d$, and $Q_T$ are weight matrices for state error, control effort, control rate change, and terminal state error, respectively. 
The control input constraints in \eqref{mpc:constraints1} are defined as follows:
\begin{equation}
     |a| \leq a_{\text{max}}, \ | r | \leq r_{\text{max}}.
    \label{mpc:constraints}
\end{equation}
To enforce collision avoidance while adhering to COLREGs rules, we define ellipse-shaped constraints in the path-based coordinate system. For left-side avoidance:
\begin{equation}
    \frac{(e- (o_e -2 o_r + m) )^2}{m^2} + \frac{(d-o_d)^2}{n^2} \geq 1,
    % \frac{(d - d_{c,i})^2}{a_i^2} + \frac{(y - y_{c,i})^2}{b_i^2} \geq 1
\end{equation}
and for right-side avoidance:
\begin{equation}
    \frac{(e- (o_e + 2 o_r - m) )^2}{m^2} + \frac{(d-o_d)^2}{n^2} \geq 1,
\end{equation}
where $(o_{d}, o_{e})$ is the obstacle's center in the path frame, and $o_r$ is the safe radius. The parameters $m$ and $n$ are the semi-major and semi-minor axes of the elliptical constraints. These constraints guide the ship to avoid the designated side in compliance with COLREGs.
An example of the MPC collision avoidance setup is shown in Fig.~\ref{fig:mpc_ex}. As depicted, the avoidance region accounts for the safe radius and ensures that the own ship maintains safe separation while avoiding the traffic ship on the correct side.

To solve the nonlinear MPC problem in real time, we employed the real-time iteration scheme from the ACADO Code Generation Toolkit \cite{houska2011acado}. The underlying nonlinear programming problem is solved using a sequential quadratic programming method, and the resulting QP subproblems are solved using the parametric active-set solver qpOASES \cite{ferreau2014qpoases}.

\bibliographystyle{IEEEtran}
\bibliography{IEEEabrv,main}
% \balance
\vspace{0cm}

\end{document}